\documentclass[a4paper,fleqn]{cas-sc}

\usepackage[numbers]{natbib}
\usepackage{caption}
\usepackage{graphicx}
\usepackage{multirow}
\usepackage{algorithm}
\usepackage{mathtools}
\usepackage[noend]{algpseudocode}
\makeatletter
\def\algbackskip{\hskip-\ALG@thistlm}

\def\tsc#1{\csdef{#1}{\textsc{\lowercase{#1}}\xspace}}
\tsc{WGM}
\tsc{QE}
\tsc{EP}
\tsc{PMS}
\tsc{BEC}
\tsc{DE}

\usepackage{fancyhdr}
\usepackage{amsmath}
\usepackage{booktabs}
\usepackage{diagbox}
\DeclareMathOperator*{\argmax}{\arg\!\max}

\usepackage{longtable}
\usepackage{lscape}
\usepackage{enumitem}
\usepackage{url}
\usepackage{hyperref}
\usepackage{longtable}
\usepackage{booktabs}

\ExplSyntaxOn
\cs_gset:Npn \__first_footerline:
  { \group_begin: \small \sffamily \__short_authors: \group_end: }
\ExplSyntaxOff

\begin{document}

\let\WriteBookmarks\relax
\def\floatpagepagefraction{1}
\def\textpagefraction{.001}
\shorttitle{Indian Spoken Language Recognition: Review}
\shortauthors{Dey et~al.}

\title [mode = title]{An Overview of Indian Spoken Language Recognition from Machine Learning Perspective}                      



\author
{Spandan Dey}[orcid=0000-0003-1770-754X]
\ead{sd21@iitkgp.ac.in}

\credit{Conceptualization of this study, Methodology, Formal analysis, Software, Writing - original draft preparation}

\address{Dept. of E\&ECE, Indian Institute of Technology Kharagpur, Kharagpur, West Bengal, India}

\author
{Md Sahidullah}[orcid=0000-0003-1770-754X]
\ead{md.sahidullah@inria.fr}

\credit{Conceptualization of this study, Methodology, Formal analysis, Writing - review \& editing}

\address{Universit\'{e} de Lorraine, CNRS, Inria, LORIA}

\author
{Goutam Saha}
\ead{gsaha@ece.iitkgp.ac.in}
\credit{Validation, Writing - review \& editing, Supervision}

\begin{abstract}
Automatic spoken language identification (LID) is a very important research field in the era of multilingual voice-command-based human-computer interaction (HCI). A front-end LID module helps to improve the performance of many speech-based applications in the multilingual scenario. India is a populous country with diverse cultures and languages. The majority of the Indian population needs to use their respective native languages for verbal interaction with machines. Therefore, the development of efficient Indian spoken language recognition systems is useful for adapting smart technologies in every section of Indian society. The field of Indian LID has started gaining momentum in the last two decades, mainly due to the development of several standard multilingual speech corpora for the Indian languages. Even though significant research progress has already been made in this field, to the best of our knowledge, there are not many attempts to analytically review them collectively. In this work, we have conducted one of the very first attempts to present a comprehensive review of the Indian spoken language recognition research field. In-depth analysis has been presented to emphasize the unique challenges of low-resource and mutual influences for developing LID systems in the Indian contexts. Several essential aspects of the Indian LID research, such as the detailed description of the available speech corpora, the major research contributions, including the earlier attempts based on statistical modeling to the recent approaches based on different neural network architectures, and the future research trends are discussed. This review work will help assess the state of the present Indian LID research by any active researcher or any research enthusiasts from related fields.
\end{abstract}


\maketitle

\section{Introduction}
In the field of \emph{artificial intelligence}, the term automatic \emph{spoken language identification} (LID) describes the ability of machines to identify the language from speech correctly. LID research aims to efficiently replicate the language discriminating human ability through computational means \cite{li2013spoken}. Due to the evolution of smart electronic gadgets, the mode of human-computer interaction (HCI) is shifting rapidly from textual typing to verbal commanding. The machines need to identify the language from the input voice command to operate on multiple languages efficiently. Therefore, for various multilingual speech processing applications, such as \emph{automatic speech recognition} (ASR)~\cite{tong2017investigation}, \emph{spoken emotion recognition} (SER)~\cite{jain2020towards}, or \emph{speaker recognition} (SR)~\cite{matvejka2017analysis}, there is a trend of using language-adapted models, that can further improve the overall recognition performance.

\thispagestyle{fancy}
\fancyhf{}
\chead{ \footnotesize Accepted for publication in ACM Transactions on Asian and Low-Resource Language Information Processing \\
DOI: 10.1145/3523179}
\renewcommand{\headrulewidth}{0pt}

According to the 23rd edition of Ethnologue, there are approximately 7,117 living languages present globally. Based on the origin of evolution, these languages are grouped into different language families. Languages that are present within a language family have evolved from the same parent language. Some of the most widely spoken language families are Indo-European, Afro-Asiatic, Sino-Tibetan, Dravidian, Austronesian, etc. Languages are governed by distinct rules in terms of syntactical, morphological, phonological, and prosodic characteristics. These language governing rules can be used as cues for individual language identification. Syntactical cues frame the set of rules in which a sentence and phrase are composed of words \cite{ling1}. Morphological cues deals with the internal structure of words \cite{morph}. Phonemes represent the basic acoustic unit for pronunciation. Generally, all the languages have a set of 15 to 50 phonemes \cite{li2013spoken}. Even if different languages share an overlapped phoneme repository, every language has some specific rules based on which phonemes can be joined as a sequence. These constraints on the legal sequence of permissible phonemes in a language are known as phonotactics. Prosodic cues represent various perceptual qualities of speech \cite{prosody}, such as intonation, rhythm, stress, etc.

We, humans, try to recognize different languages based on these linguistic cues present at different levels. With adequate knowledge of a language, human intelligence is undoubtedly the best language recognizer than the trained LID machines~\cite{ambikairajah2011language}. Even if someone is not familiar with a particular language, based on the different linguistic cues, humans can approximately provide a subjective judgment about that unknown language by correlating it to a similar-sounding language. There are several practical applications where a human operator is needed with multilingual skills. For example, in call center scenarios, a human operator often needs to route the telephone calls to the proper representative based on the query's input language. In certain security and surveillance applications, knowledge of multiple languages is needed. Deploying a trained LID machine over a human operator is always more effective in these scenarios.  Humans can be trained in a limited number of different languages for accurate classification. Almost $40 \%$ of the world population is monolingual, $43 \%$ are bilingual, whereas only $13\%$ of the human population is trilingual~\footnote{\url{http://ilanguages.org/}}. Because of this reason, if the number of languages needed to be recognized is sufficiently large, it will be increasingly difficult for humans to perform language identification tasks accurately. Moreover, training human operators in multiple languages is a time-consuming and challenging task. 

Another increasingly popular application for automatic spoken language recognition is for multilingual verbal interaction with smart devices. There are various popular voice assistant services, such as Apple's Siri, Amazon's Alexa, Google voice assistant, etc., which share more than $40\%$ of the global voice assistant market~\footnote{\url{https://www.marketresearchfuture.com/reports/voice-assistant-market-4003}}. As per~\footnote{\url{https://www.forbesindia.com/blog/technology/voice-tech-industrys-next-big-platform-shift/}}, by 2020, almost $50\%$ of the web searches are already in voice commands rather than typing, and the numbers are expected to grow more. \emph{Internet of things} (IoT) based smart gadgets are also being enabled with various speech processing applications for verbal interaction. A front-end LID block is essential for reliable performance across users of different languages for all of these voice-enabled applications. For example, in the case of speech recognition systems used in voice assistants, individual language-based models improve overall recognition accuracy~\cite{hemakumar2013speech}. Smart speaker devices can also be improved by developing individual language-specific speech synthesis models. Similarly, training different classifiers based on individual languages is helpful for speaker verification systems or emotion recognition systems. By automatically detecting the language from the input speech, the smart devices can change the mode of operation dedicated to a particular language, improving the user experience for voice-enabled smart devices under multilingual conditions.

\begin{figure}[t]
\vspace{-.1cm}
    \centering
    \includegraphics[trim={0cm 0cm 0cm   0cm},clip,width=.8\textwidth]{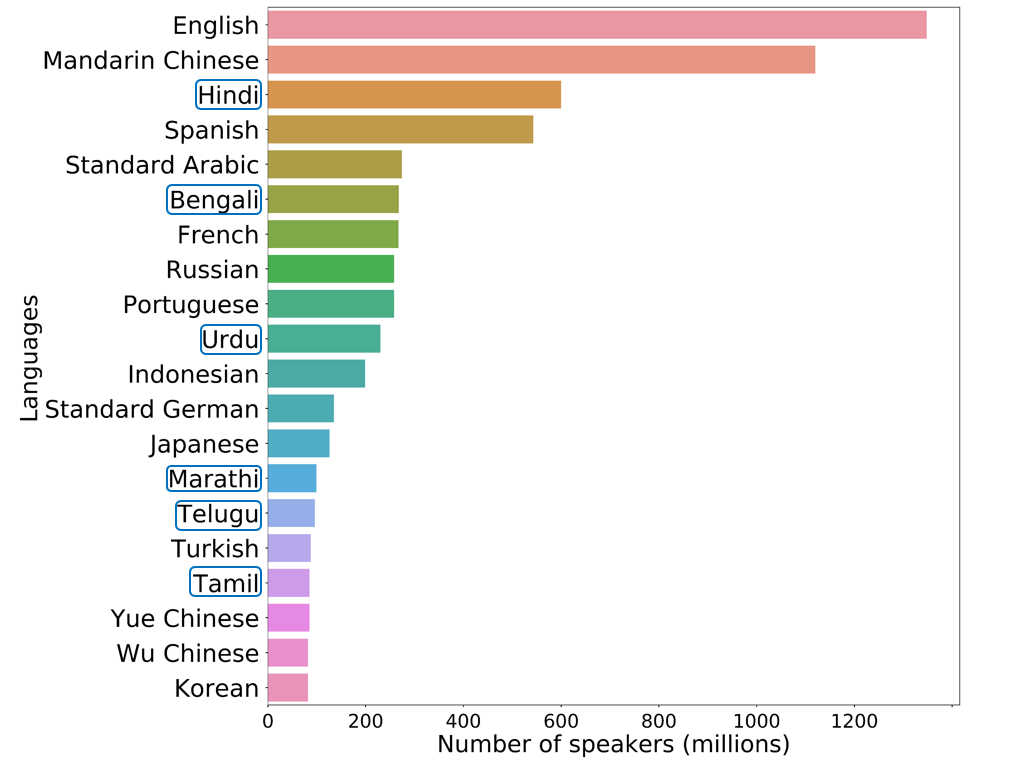}
    \vspace{-.1cm}
    \caption{Number of speakers (in millions) for the twenty most widely spoken languages in the world with the Indian languages marked in the boxes.}
    \label{fig:horiz_lang}
    \vspace{-.7cm}
\end{figure}

India is the second-most populous country in the world with a total population of more than 1.3 Billion\footnote{\url{https://data.worldbank.org/}}. This massive population is also culturally very diverse and has different native languages. There are more than 1500 languages present in India~\footnote{\url{https://censusindia.gov.in/2011Census/Language_MTs.html}}. The Indian constitution has given 22 languages the official status, and each of them has almost more than one million native speakers. In Fig.~\ref{fig:horiz_lang} twenty most widely spoken languages in the world~\footnote{\url{https://www.ethnologue.com/ethnoblog/gary-simons/welcome-24th-edition}} are shown with the number of speakers in millions. Of these twenty languages, six languages (Hindi, Bengali, Urdu, Marathi, Telugu, and Tamil) are mainly spoken in India and South Asia. In the last decade, a significant portion of the Indian population has become quite familiar with several smart electronic devices. However, the majority of the Indian population is more comfortable with their native languages rather than English or other global languages~\cite{plauche2006speech}. Even if someone is comfortable verbally interacting with the smart devices in Indian English, issues related to different accents often arise. If these smart devices can be operated by speech commands, especially in the local languages, the entire population of India can use them with ease. For that purpose, researchers are trying to develop various speech processing applications, such as automatic speech recognition~\cite{kumar2005development,hemakumar2013speech,singh2019asroil,fathima2018tdnn}, speech synthesis~\cite{panda2020survey,baljekar2018investigation}, speaker recognition~\cite{haris2012multivariability}, etc. for individual Indian languages. While developing multilingual technologies for the Indian context, a front-end Indian spoken language classifier block is very important. The LID block is responsible for automatically detecting the input language and switching the mode of operation dedicated to the detected language. Therefore, developing efficient LID technologies for Indian languages is an important field of research for the technological advancement of more than one billion population.

For more than three decades, the research field of spoken language recognition has been developing. For many languages, such as English, Mandarin, and French, the state of LID has become mature enough for satisfactory deployment in real-world scenarios. However, compared to that, the progress for LID research in Indian languages was initially very limited. The main challenge for developing efficient Indian LID systems was due to the unavailability of large, standard multilingual speech corpora for Indian languages~\cite{maity2012iitkgp}. During the last decade, due to the development of various standard speech corpora for the Indian languages, it has been possible to conduct extensive research in this Indian LID field~\cite{shrishrimal2012indian,kiruthiga2012design}. Researchers have also identified some unique challenges for developing LID systems for Indian scenarios. For example, many of the widely spoken Indian languages are still \emph{low-resourced} in terms of the availability of speech corpora~\cite{kurian2015review}. Even within the same language, the dialect changes widely. The majority of the Indian languages belong to only two language families and share a common phoneme repository~\cite{sengupta2016identification,jothilakshmi2012hierarchical}. The mutual influence among the languages also made the development of accurate discrimination of individual languages more challenging~\cite{sengupta2015study}. Researchers are actively trying to address these challenging issues for developing efficient Indian spoken language identification systems. As a result, a significant amount of research has been made to develop Indian LID systems from different perspectives. Various levels of speech features have been explored for discriminating the Indian languages, such as acoustic features~\cite{koolagudi2012identification,nandi2015implicit,dutta2018language,siddhartha2020language}, phonotactic~\cite{mohanty2011phonotactic}, prosodic~\cite{reddy2013identification,bhanja2019pre}, bottleneck~\cite{das2020bottleneck}, fused features~\cite{das2020hybrid,deshwal2020language}, etc. From a classifier perspective, there were already some attempts based on generative models~\cite{manwani2007spoken,kumar2015significance,reddy2013identification,nandi2015implicit}. In the last few years, several deep neural network (DNN) architectures have been extensively utilized for improving LID performance~\cite{mounika2016investigation,veera2018combining,bhanja2019deep,mandava2019attention,mandava2019investigation}. 

For spoken language recognition, from in-general perspectives, there are several comprehensive review papers~\cite{ambikairajah2011language,li2013spoken} and collaborative research works~\cite{singer2003acoustic,zissman1996comparison}. However, most of these works were almost a decade ago. The recent trends of the LID research are needed to be summarized. For the languages of India and South Asia, there are few prior attempts of providing collective information about the development of LID systems. The review work by Aarti et al. (2018)~\cite{aarti2018spoken} discussed several language-dependent features and databases for the Indian languages. Shrishrimal et al. (2012) surveyed the existing speech corpora available for the Indian languages. Some of the research works, such as~\cite{mounika2016investigation,reddy2013fusion}, along with their technical contributions, provided some additional discussions for the existing literature. However, all of these prior works had focused on individual aspects of the LID research. We find that for the Indian languages, there is yet no extensive review work that covers the entire spectrum of topics in detail. It motivates us to present one of the first extensive analytical reviews of the major research works to address the Indian spoken language recognition task. In this work, We have also discussed the fundamentals of Indian LID systems, the unique challenges for language recognition in the Indian context, and the description of the available standard speech corpus for India LID. The unsolved open issues, present trends, and future research directions are also discussed. The goal is to provide a complete overview of the present state of the research for the developers and the research enthusiasts of the Indian spoken language recognition problem. The major contribution of this work is listed as follows:

\begin{itemize}
    \item To the best of our knowledge, this is one of the first attempt to collectively review the all of the major research contributions made in Indian spoken language recognition research.
    
    \item From a linguistic perspective, we have discussed the unique challenges faced, especially for developing LID systems in the Indian context. These challenges can help modify the existing LID systems for optimum performance in the Indian scenario.
    
    \item Even for the global language recognition scenario, there is already a significant amount of time has passed since the last known complete review work. Keeping this fact in mind, an overview of the recent advances in the overall language recognition work is also presented. Special attention is given to reviewing the recent neural network-based research attempts. It will help the readers get a comprehensive insight into the recent advances in LID research for global languages.
    
    \item Finally, we discuss the unsolved open challenges in the Indian language recognition research, followed by our analysis of the potential future research directions.
\end{itemize}

The rest of the paper is presented as follows: Section \ref{sec:2} elaborates the fundamentals of language recognition systems. In Section \ref{sec:3}, the requirements and challenges of Indian spoken language recognition systems are discussed. In Section \ref{sec:4}, a detailed description of the developed Indian LID speech corpora is presented. A review of the major research progresses for Indian LID is carried out in Section \ref{sec:5}. Section \ref{sec:6} summarizes the review work with the open challenges and potential future research trends. We have concluded this paper in Section \ref{sec:7}.

\section{Fundamentals of Spoken Language Recognition}
\label{sec:2}
Automatic spoken language identification (LID) can be formulated as a pattern recognition problem. LID system consists of a front-end feature extraction unit followed by a classifier backend. The front end efficiently extracts language discriminating information from raw speech waveform by reducing other redundant information. These extracted features are then used as the input of the classifier block.  

In the front-end feature extraction block, at first, the speech waveform is segmented into frames. Framing is done by multiplying the speech waveform by successive overlapping windows \cite{benesty2007springer}. Then, for each of the frames, following certain parameterization rules, feature $\mathbf{x} \in \mathbb{R}^{N_x}$ is computed. Here, $N_x$ is called the feature dimension. If the sampling rate of the speech file is $F_s$ and the time duration of each speech frame is $t$ seconds (s), then the total number of samples in each speech frame can be calculated as $N_{frame}=F_s*t$. In the feature space, raw speech is transformed into a much more compact representation as, $N_x << N_{frame}$. The total number of frames ($T$) for a speech segment depends on the overlap between successive framing windows. After the feature extraction, for each speech utterances, a set of feature vectors $\mathbf{X}=\{\mathbf{x_1}, \mathbf{x_2}, ..., \mathbf{x_T}\}$ is generated, where $\mathbf{X} \in \mathbb{R}^{N_x \times T}$. These feature vectors are then fed to the classifier block.

Classifiers can be categorized into generative models and discriminative models based on the manner they learn the discriminating cues of the target classes from the input feature set \cite{bishop2006pattern}. Generative models learn the distribution of feature space for each languages class during training \cite{bishop2006pattern}. At the end of the training, individual language models, $\lambda_i$ are learned, where $i=\{1, 2, ..., L\}$ and $L$ denotes the number of languages to be recognized. A language model is defined as the set of parameters for estimating the distribution of the feature space of a particular language. During the testing phase, the feature vector of the test speech segment is fed to each of the $L$ language models. Each of the language models $\lambda_i$ produces the posterior probability $P(\lambda_i|\mathbf{X})$, depicting the chance of occurrence of a particular language in the test segment provided the input feature $\mathbf{X}$. The predicted language $\hat{L}=L_p$ is computed by \emph{the maximum a posterior probability} (MAP) criteria \cite{ambikairajah2011language}:

\begin{equation}
p=\argmax_{1\leq i\leq L}P(\lambda_i|\mathbf{X})  
\label{eq: map}
\end{equation}
where, p=$1, 2,\ldots, L$. Again by further expanding Eq.~\ref{eq: map} using Bayes' rule:
\begin{equation}
    p = \argmax_{1\leq i\leq L}\frac{P(\mathbf{X}|\lambda_i)P(\lambda_i)}{P(\mathbf{X})}
    \label{eq:expand map}
\end{equation}
Here, $P(\lambda_i)$ denotes the prior probability of the $i^{th}$ language class. $P(\mathbf{X})$ term is independent of the language class $i$. Hence in Eq.~\ref{eq:expand map}, it can be ignored by treating as a constant. If we assume that the chance of the occurrence of each language is equally likely, then the MAP criteria of prediction in Eq.~\ref{eq: map} is simplified into the \emph{maximum likelihood} (ML) criteria \cite{bishop2006pattern}. 
\begin{equation}
    p =\argmax_{1\leq i\leq L}P(\mathbf{X}|\lambda_i)
    \label{eq:ml}
\end{equation}

Discriminative classifiers are not trained to generalize the distribution of the individual classes. Rather, it learns a set of weights $\mathbf{w}$, which is used to define the decision boundary function $g(\mathbf{w},\mathbf{X})$ among the classes. For $L$ language classes, the model output can be denoted as $\mathbf{s} \in \mathbb{R}^{L}$. Generally, $\mathbf{s}$ is then transformed to a \emph{softmax}~\cite{goodfellow2016deep} score vector:
\begin{equation}
    \sigma(s_i)=\frac{e^{s_i}}{\sum_{j=1}^Le^{s_j} } \mbox{  for i = 1, 2, \ldots, L}
    \label{eq: CE}
\end{equation}
The softmax score is then used to calculate a loss function. The loss function measures how close the predicted and true values are. Usually for multi-class classification, \emph{categorical cross-entropy} loss \cite{goodfellow2016deep} is calculated:
\begin{equation}
    CE=-\sum_{i=1}^{L}t_i \log \sigma(s_i)
\end{equation}
Here, $t_i$ denotes the true label (either 0 or 1) of the $i^{th}$ language for a speech segment. The weights $\mathbf{w}$ are learned by optimizing this loss function. During testing, the feature vector extracted for the test segment is fed to the model. Similar to the training phase, the model outputs a softmax score vector $\mathbf{s}$. Finally, the predicted language class $\hat{L}=L_j$ can be expressed as:
\begin{equation}
    j= \argmax {\mathbf{s}}
    \label{eq: discriminative classifiers}
\end{equation} where j=$1, 2,\ldots, L$.

\subsection{Description of different language discriminating features present in speech}
\begin{figure}[htbp]
    \vspace{-.3cm}
    \centering
    \includegraphics[trim={.7cm 0cm .7cm 0cm ,clip},width=.7\linewidth]{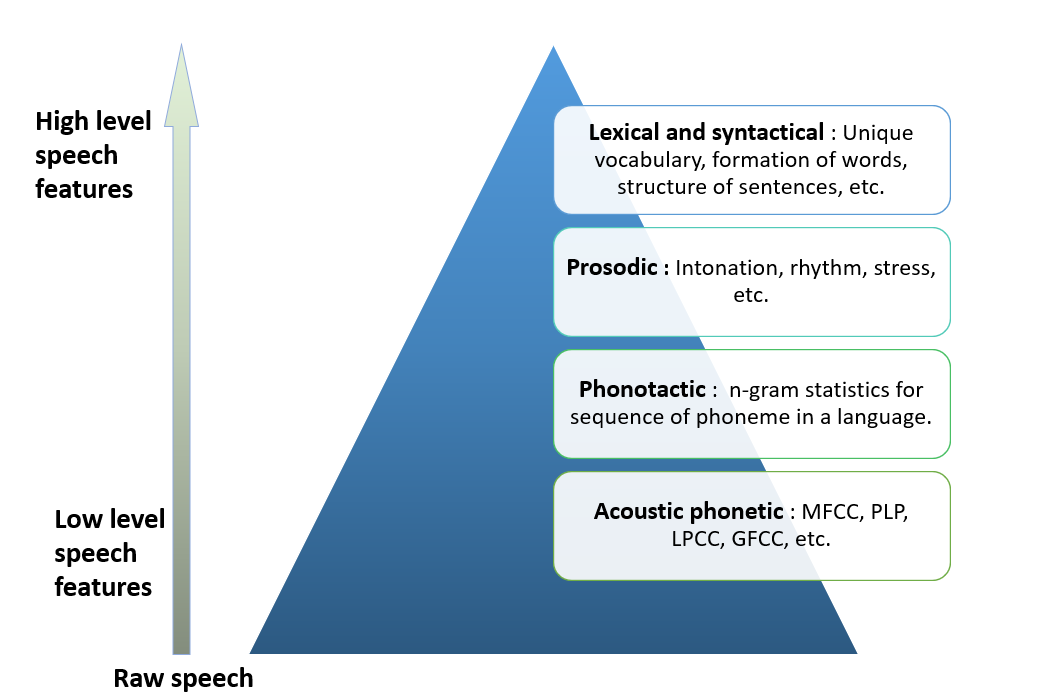}
    \vspace{-.3cm}
    \caption{Hierarchical representation of different language discriminating information present in speech.}
    \label{fig:hierarchical}
    \vspace{-.4cm}
\end{figure}
In speech signal, along with the language information, several other information is also present, such as speaker identity, speaker emotions, channel characteristics, and background noise~\cite{li2013spoken}. For building LID systems, the front end feature extraction transforms the speech signal into a compressed representation by removing the redundancies and retaining the necessary language discriminating cues \cite{li2013spoken}. These cues are present in speech at different levels, extracted by different levels of speech features. Lower level speech features, such as acoustic, phonotactic, prosody, are directly extracted from raw speech waveform. High-level features, such as lexical and syntactical features, contain more language-discriminating information.  However, they can not be extracted directly from raw speech \cite{ambikairajah2011language}. Therefore, in most LID literature, research progress mainly focuses on exploring low-level features. In Fig.~\ref{fig:hierarchical}, the multi-level language discriminating speech features are shown in the hierarchical order.

\subsubsection{Acoustic-phonetic features:}
These features explore the amplitude, frequency, and phase information of speech waveform. Due to the ease of extraction, acoustic features are also used to formulate higher-level speech features. \emph{Mel frequency cepstral coefficients} (MFCC), \emph{perceptual linear prediction} (PLP), \emph{linear prediction cepstral coefficients} (LPCC), \emph{contant-Q cepstral coefficient} (CQCC), \emph{gammatone frequency cepstral coefficients} (GFCC) are the most commonly used acoustic features. The majority of these features are derived from the magnitude spectrum of \emph{short time Fourier transform} (STFT). Similarly, using the phase part of STFT, several acoustic features are used for LID tasks \cite{dutta2018language}. The frame-wise acoustic features are called static features. In literature, after extraction of static features, contextual information from adjacent frames are also concatenated by $\Delta$, $\Delta^2$~\cite{ambikairajah2011language}, and \emph{shifted delta coefficients} (SDC)~\cite{wang2012shifted,torres2002approaches} features. SDC features are widely used in the LID literature. They are shown to outperform the $\Delta-\Delta^2$ features for LID task~\cite{torres2002approaches} because of their ability to span a larger number of adjacent frames for collecting the contextual information~\cite{bielefeld1994language}. The computation of the SDC feature is shown in Eq.~\ref{eq:sdc}.
\begin{equation}
    \Delta c(t+iP)=c(t+iP+d)-c(t+iP-d)
    \label{eq:sdc}
\end{equation}
Here, $i \leq 0<k$. Four parameters $(N, d, P, k)$ are used for SDC computation. $N$ is the dimension of the static features for each speech frame, $d$ denotes the number of frames advance and delay to compute the delta feature, $k$ is the number of blocks whose delta coefficients are concatenated to form the final feature vector, and $P$ is the frameshift between the blocks. Thus, for each frame, SDC computes $kN$ coefficients for context, whereas the $\Delta-\Delta^2$ uses $2N$ coefficient for context.

\begin{figure}[!t]
    \centering
    \includegraphics[trim={2.5cm 0.2cm 3cm  1cm},clip,width=.94\textwidth]{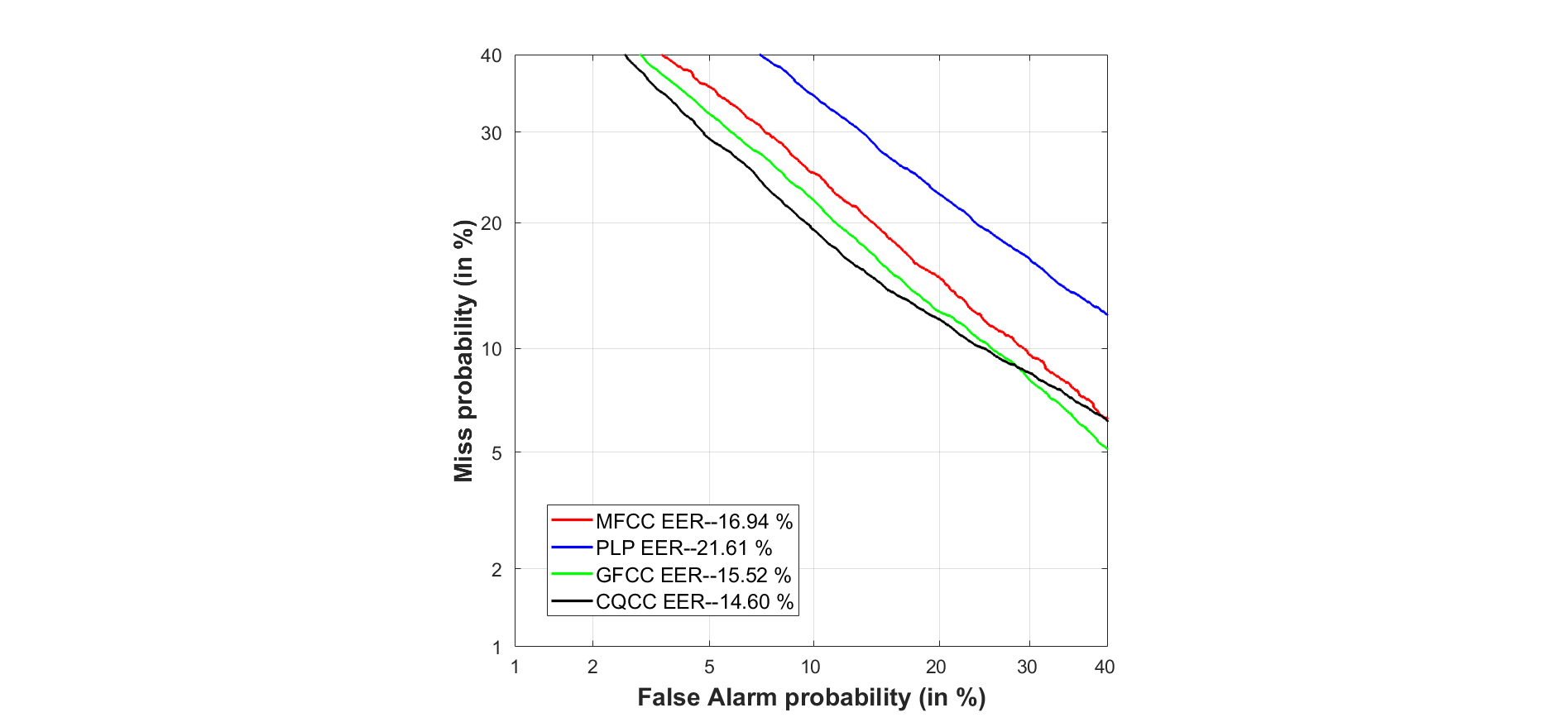}
    \vspace{-.35cm}
    \caption{Comparison of language recognition performance of four acoustic features using the DET curve.}
    \label{fig:det}
    \vspace{-.65cm}
\end{figure}

In Fig.~\ref{fig:det}, the comparative LID performance of four acoustic features are shown in terms of \emph{detection error trade-off} (DET) curve and \emph{equal error rate} (EER) (see Section~\ref{subsec:metric}). Eight of the most widely spoken Indian languages, Hindi, Bengali, Marathi, Telugu, Tamil, Gujarati, Urdu, and Punjabi, are selected from the IIITH-ILSC database \cite{vuddagiri2018iiith}. For all the acoustic features, 13-dimensional static cepstral coefficients are used. For classification, \emph{TDNN based x-vector} architecture~\cite{snyder2018x} is used. This architecture contains five TDNN layers followed by a statistics pooling layer. The TDNN layers incorporate a context of 15 frames. After the pooling layer, two fully connected layers are used. The experiment is conducted using the PyTorch~\cite{paszke2019pytorch} library with NVIDIA GTX 1050Ti GPU. We have used batch size of 32. AdamW~\cite{loshchilov2019decoupled} optimizer is used. We train the architecture for 20 epochs and use an early-stopping criterion of 3 epochs based on the validation loss. The DET curve shows that all the acoustic features are able to classify the eight languages decently in a comparable range.

\subsubsection{Phonotactic features:}
The phonological rules for combining various phonemes differ among languages. Phonotactic features explore this uniqueness of permissible phone sequences in a language. Different languages may have overlapping sets of phoneme repositories, but some particular phone sequences may be invalid or rarely used in one language, whereas the same phone sequence can be widely used in the other \cite{ambikairajah2011language}. Phonotactic features are derived using the n-gram phone statistics. Different approaches have been applied for classifying languages using phoneme statistics \cite{zissman1996comparison},~\cite{diez2012use},~\cite{kukanov2020maximal}.
\subsubsection{Prosodic feature:}
Prosodic features represent the perceptual qualities of the speech signal in a quantitative manner~\cite{reddy2013identification}. Some of the commonly explored prosodic ques in LID research are intonation, rhythm, and loudness. Features extracted from pitch contour statistics can effectively formulate intonation. Whereas the duration of syllables, fraction of voiced to unvoiced syllables, duration contour statistics are used to explore the rhythmic nature of speech. Loudness of speech can be parameterized using log energy, intensity contour characteristics. Prosodic features are shown to be more noise-robust as compared to acoustic features~\cite{ng2013spoken}. 

\begin{figure}[h]
    \centering
    \vspace{-.33cm}
    \includegraphics[trim= .6cm .5cm 1cm .4cm,clip,width=\textwidth]{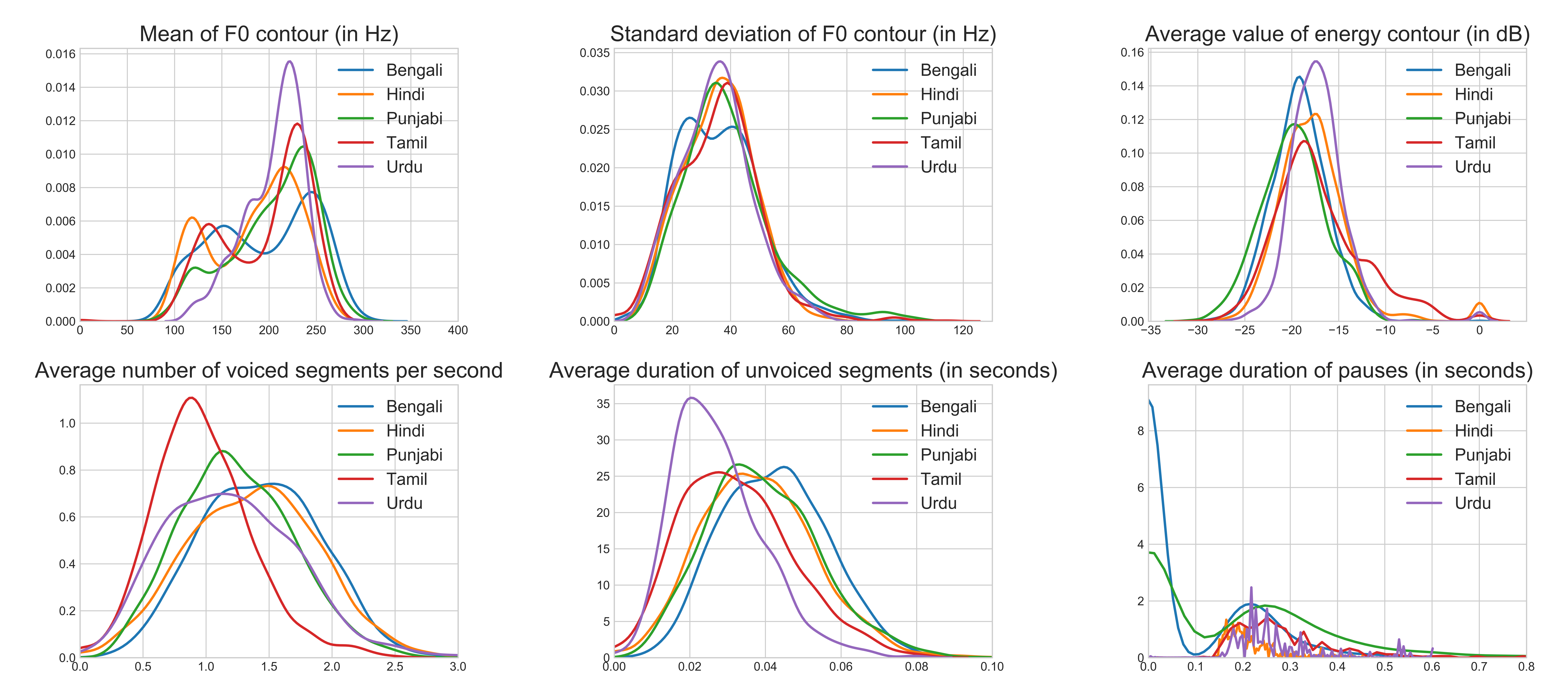}
    \vspace{-.8cm}
    \caption{Language discriminating capabilities of various prosody features.}
    \label{fig:prosody_dist}
    \vspace{-.55cm}
\end{figure}

In Fig.~\ref{fig:prosody_dist}, we have shown the language discriminating capabilities of several intonation, duration, and energy-based prosody features. We have selected five Indian languages, Bengali, Hindi, Punjabi, Tamil, and Urdu, from the IIITH-ILSC database~\cite{vuddagiri2018iiith}. For each of the languages, we have computed the mean prosodic feature values across the utterances. After that, we plot the distribution of these mean values. For each prosodic feature, there are five distribution plots corresponding to the five target classes. For a particular prosodic feature, the more the inter-class distributions differ, the better is the language discriminating capability. The mean $F_0$ distributions are bimodal. The lower mode corresponds to the male speakers, and the higher mode corresponds to the female speakers. We can observe a gradual increase in the frequency for the higher mode value for Urdu, Tamil, Punjabi, and Bengali, respectively. Hindi has the lowest frequency for both two modes, indicating better suitability for recognition. For the standard deviation of $F_0$ plots, Bengali can be better classified from the rest. Apart from Bengali, all other languages show similar uni-modal nature. Tamil data has more utterances with higher average energy values and a lesser number of voiced segments per second. Similarly, Urdu utterances are distinctly recognized using the average duration of unvoiced segments per second. Bengali and Punjabi utterances show the tendency of having more utterances with lower averaged pause duration indicating better recognition with these prosodic cues.
\subsubsection{Lexical and syntactical features:}
These are the high-level features. Lexical features explore the unique rules of word formation in a language. Syntactical features use the different rules for sentence formation using the words. Often, \emph{large vocabulary speech recognizer} (LVSR), trained on low-level features, are required to fetch this information. For many low-resourced languages, developing LVSR is not trivial. Therefore, high-level features are utilized less compared to the low-level features for the LID task.

\subsubsection{Bottleneck features:}
In the last few years, ASR bottleneck features are used efficiently for the LID task~\cite{matejka2014neural},~\cite{richardson2015deep}, ~\cite{fer2015multilingual},~\cite{mclaren2016exploring},~\cite{lozano2017analysis}. From the transcripts, first, a large ASR classifier is trained for the corresponding languages. Then the embeddings from these trained ASR models are extracted and used for the LID task. The languages for the LID do not require to have their transcripts in this approach. Bottleneck features contain complementary information as compared to the acoustic feature. Hence if the bottleneck information is utilized with the acoustic systems, the overall LID performance and robustness improve.

\subsubsection{Feature post-processing:}
\begin{figure}[!h]
    \centering
    \vspace{-.3cm}
    \includegraphics[trim={.5cm 0cm .5cm 0.1cm ,clip}, width=.65\linewidth]{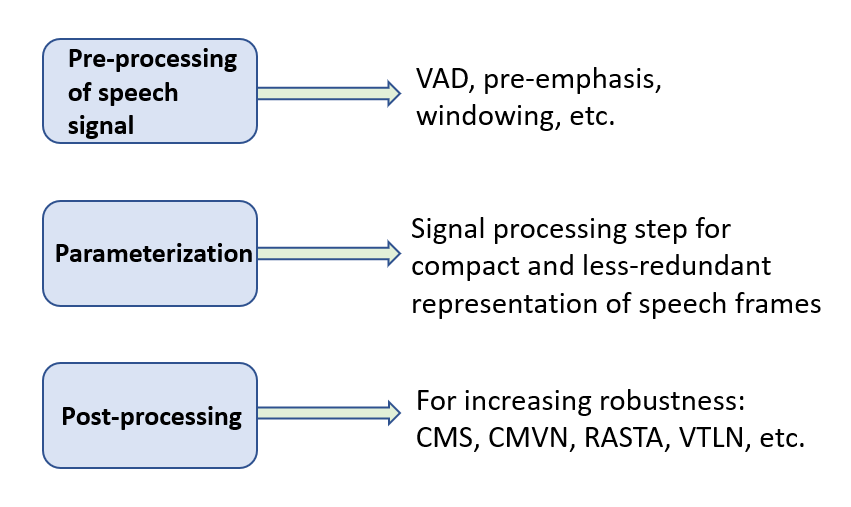}
    \vspace{-.55cm}
    \caption{Different steps for feature extraction from raw speech signals.}
    \label{fig:feat_extract}
    \vspace{-.45cm}
\end{figure}

Before the feature extraction, during the pre-processing steps, silence is removed, high frequency components are emphasized and windowing is performed to transform speech signals into overlapping frames. After feature extraction, to make the features robust against background noise and channel variations, post processing steps such as \emph{cepstral mean subtraction} (CMS) \cite{pelecanos2001feature}, \emph{cepstral mean and variance normalization} (CMVN) \cite{kalinli2019parametric}, \emph{RASTA} filtering~\cite{hermansky1994rasta}, \emph{vocal tract length normalization} (VTLN)~\cite{eide1996parametric} are applied. Recently, trainable feature post-processing techniques are being efficiently applied for several speech based classification tasks. In the trainable configuration, parameters for the post-processing techniques can be jointly learnt with the language training. For example, \emph{parametric cepstral mean normalization}~\cite{kalinli2019parametric} and \emph{per-channel energy normalization} (PCEN)~\cite{wang2017trainable,lostanlen2018per} are shown to be more robust than the conventional feature post-processing for speech recognition. The different steps for feature extraction is shown in Fig.~\ref{fig:feat_extract}. 

Following the existing literature, high-level features, although they contain more language discriminating cues, are not commonly used for the LID task. Rather, due to the simple extraction procedure, acoustic features are most commonly used. The usage of phonotactic features also requires transcribed corpus for the language, which is often not available for the low-resourced languages. Bottleneck features from a pre-trained ASR model are also a preferred choice by researchers~\cite{silnova2018but}. The languages within a language family may also share common phoneme repository and phonotactic statistics~\cite{mounika2016investigation}. In such scenarios, phoneme-based LID may be challenging. Prosody can also be fused with acoustic~\cite{reddy2013identification, bhanja2019pre} or phonotactic systems~\cite{ng2013spoken} to provide complementary language discriminating information. Acoustic features extract information from the frequency domain. Time domain speech features~\cite{benesty2007springer}, such as amplitude envelops, short-time average \emph{zero-crossing-rate} (ZCR), \emph{short-time average energy}, \emph{short-time averaged magnitude difference} and \emph{short-time auto-correlations} can also be used as a complementary source of information for the LID task. We find that the optimal feature for the LID task is selected intuitively based on problem specification, and it is an open area of research.

\subsection{Description of different modeling techniques for LID task}
The research progress of automatic spoken language recognition has come down a long way. Apart from exploring different features, various modeling techniques have been successfully applied for LID tasks. In the initial phase of language recognition research, various modeling techniques, such as, \emph{hidden Markov model} (HMM) \cite{house1977toward} and \emph{vector quantization} (VQ) \cite{sugiyama1991automatic} were applied. Initially, LID models based on HMM were an intensive area of research because of the ability of the HMM models to capture contextual information. 

Classifier models based on phone recognition and phonotactic n-gram statistics were also a popular approach for language recognition \cite{zissman1996comparison}. \emph{phone recognizer followed by language modeling} (PRLM) \cite{zissman1994automatic} technique used a front end phone recognizer for language classification. A further extension of this model was made in \emph{parallel phone recognition followed by language modeling} (PPRLM) approach \citestyle{hazen1997segment,zissman1994automatic}. Instead of using only a single phone recognizer as the front end, several independent phone recognizers were trained in different languages~\cite{zissman1996comparison}. PPRLM technique was shown to outperform the PRLM based LID systems. Another technique used for phonotactic modeling was \emph{parallel phone recognizer} (PPR) based language modeling~\cite{zissman1996comparison}. 

The phonotactic-based techniques were suitable only for the languages that have phonetically transcribed speech corpora. Due to several parallel subsystems, the PPRLM technique was computationally slow. The phonotactic systems were also not very robust against noise. A generative classification approach, \emph{Gaussian mixture model} (GMM), was used successfully for speaker recognition task~\cite{reynolds1995robust} and was later applied efficiently for language recognition purpose. Let, the LID system classify $L$ number of languages. Let, the training feature vector for the $l^{th}$ language class is denoted as $\mathbf{X_l}=[\mathbf{x}_{l_1}, \mathbf{x}_{l_2}, \cdots, \mathbf{x}_{l_T}] \in \mathbb{R}^{(d \times T)}$, where, $l=[1, 2, \cdots, L]$, $d$ is the input feature dimension and $T$ is the number of time frames. In GMM, the feature space distribution for each language class $l$ is modeled:
\begin{equation}
    P(\mathbf{x}|\lambda_l)=\sum_{m=1}^M w_m b_m(\mathbf{x})
\end{equation}
where, $m=1:M$ is the mixture coefficient denoting each of the $M$ number of multi-variate Gaussian distribution ($b_m$), used to model the feature space: 
\begin{equation}
    b_m(\mathbf{x})=\frac{1}{(2\pi)^{(d/2)}|\mathbf{\Sigma}|^{(1/2)}}e^{[(\mathbf{x}-\mathbf{\mu})'\mathbf{\Sigma}^{-1}(\mathbf{x}-\mathbf{\mu})]}
\end{equation}
During the training process, the GMM parameters $\lambda=\{w_m, \mathbf{\mu}_m, \mathbf{\Sigma}_m\}$ are learned for $m=1:M$. During testing, the feature vector for any test utterance $\mathbf{X}$ is presented to each of the $L$ GMM models, and corresponding likelihood scores are computed:
\begin{equation}
    P(\mathbf{X}|\lambda_l)=\prod_{t=1}^T P(\mathbf{x}_t|\lambda_l)
\end{equation}
Following the ML criteria presented in Eq.~\ref{eq:ml}, the language prediction is made. Unlike n-gram approaches, it did not require transcribed corpora. Therefore, this approach could be used for a large number of languages which do not even have transcription. GMM was further improved by using \emph{universal background model} (UBM)~\cite{reynolds2000speaker}. In the UBM approach, first a GMM model is trained with usually larger mixture coefficients and sampling training data from several language classes. Then, using the corresponding training data for each of the target-language classes, the UBM parameters are further tuned to generate the language-specific GMM models by MAP adaptation. Using this approach, the language-specific GMM parameters share a tighter coupling which improves robustness against feature perturbations due to unseen acoustic events~\cite{kumar2015significance}. The classification performance though generally saturates with the increase of Gaussian mixture components if training data is not large enough~\cite{verma2013indian}. However, with the availability of more training data, later on, several other discriminative approaches outperformed GMM-based classifiers.

\emph{Support vector machine} (SVM) was efficiently applied for language recognition in \cite{campbell2004language}. In \cite{zhai2006discriminatively}, phonotactic features were used with SVM as a modeling technique for language recognition. SVM based models in \cite{castaldo2007language}, outperformed the GMM based LID models. SVM is a discriminative model. The input feature space is not restricted in SVM; it can be mapped into higher dimensions such that a hyper-plane can maximally separate the classes. Therefore, the generalization with SVM models is generally better. For more complex classification problems, however, machine learning literature shows neural networks to perform better than the SVMs. 

The \emph{i-vector} approach was also used as feature extractor for language recognition research~\cite{dehak2011language}, \cite{ferrer2015study}, \cite{sizov2017direct}, \cite{padi2019attention}. This technique was initially used in \cite{dehak2010front}, for the speaker recognition task. First, by pooling the available development data, a UBM model $P(\mathbf{x}|\lambda)$ is trained with $\lambda=\{w_m,\mathbf{\mu}_m, \mathbf{\Sigma}_m\}$ and $m=1:M$. Here, $\mathbf{\mu} \in \mathbb{R}^d$ is the UBM mean vector for the $m^{th}$ component and $\mathbf{\Sigma} \in \mathbb{R}^{(d \times d)}$ is the UBM covariance vector for the $m^{th}$ component, $d$ is the feature dimension. For each component, the corresponding zero and centered first-order statistics are aggregated over all time frames of the utterance as:
\begin{equation}
 N_m=\sum_{t} P(\mathbf{x}_t|\lambda)   
 \end{equation}
\begin{equation}
F_m=\sum_{t} P(\mathbf{x}_t|\lambda)(\mathbf{x}_t-\mu_m)    
\end{equation}
UBM supervectors $N \in \mathbb{R}^{(Md \times Md)}$ and $F\in \mathbb{R}^{Md}$ are formed by stacking the $N_m$ and $F_m$ respectively for each M Gaussian components. The zero-order statistics $\mathbf{N}$ is represented by a block diagonal matrix with $M$
diagonal $d \times d$ blocks. The i-vector approach then estimates a \emph{total variability} matrix $\mathbf{T} \in \mathbb{R}^{(Md \times c)}$ such that:
\begin{equation}
    \mathbf{N}^{-1}F=\mathbf{T}\mathbf{w}
\end{equation}
Here, $\mathbf{w} \in \mathbb{R}^{c}$. $\mathbf{T}$ is shared for all language classes, and it captures all the variability of sessions, speakers, and languages in the training data. $\mathbf{T}$ maps the information captured in $F$ into a much smaller vector $\mathbf{w}$ such that $Md>c$. The $\mathbf{w}$ is a fixed dimension vector for each utterance. It is called the \emph{i-vector} and extracted as:
\begin{equation}
    \mathbf{w}=(\mathbf{I}+\mathbf{T}'\mathbf{\Sigma_F}^{-1}\mathbf{N}\mathbf{T})^{-1}\mathbf{T}'\mathbf{\Sigma_F}^{-1}F
\end{equation}
where, $\mathbf{\Sigma_F} \in \mathbb{R}^{(Md \times Md)}$ is the diagonal covariance matrix for $F$.
One main advantage of the i-vector is the fixed dimension of the feature vector, which is independent of the utterance duration. Due to this, classification performance can be significantly improved by various embedding post-processing techniques (centering, whitening, length normalization, etc.) followed by simple classifier backends, such as \emph{probabilistic linear discriminant analysis} (PLDA), cosine similarity, and logistic regression. The variations in posterior probabilities of prediction increases if the training data is not sufficient~\cite{gonzalez2015frame}. Computation latency is also a crucial issue for real-time applications~\cite{gonzalez2015frame}.
\begin{figure}
    \centering
    \includegraphics[trim=.15cm 0.4cm .1cm 0.1cm,clip,width=.9\textwidth]{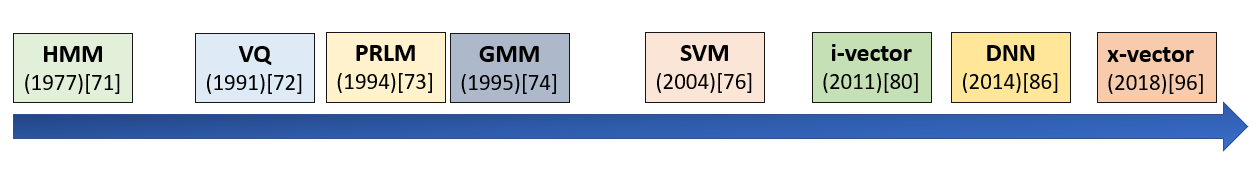}
    \vspace{-.2cm}
    \caption{Summary and timeline of the various modeling techniques applied for LID research.}
    \vspace{-.6cm}
    \label{fig:timeline}
    \vspace{-.15cm}
\end{figure}


In the last few years, several neural network based approaches have been developed for LID tasks. These neural network based approaches outperform GMM and i-vector LID systems, especially in the presence of a large amount of training data~\cite{lopez2014automatic}. In \cite{gonzalez2015frame}, the authors showed that for language recognition with short utterance duration, DNN based classifiers significantly outperformed the i-vector based models. These approaches directly applied DNN classifiers as an end-to-end modeling technique. DNN model trained for speech recognition is also used to extract bottleneck features for the following LID classifier in \cite{richardson2015deep}. \emph{Convolutional neural network} (CNN) was also explored in various research works for language recognition,\cite{montavon2009deep, lei2014application}. LID experiment was also carried out with sequential neural networks models, such as, \emph{recurrent neural network} (RNN) \cite{geng2016end}, \emph{long short term model} (LSTM) \cite{gonzalez2014automatic,zazo2016evaluation}, \emph{bi-directional LSTM}~\cite{fernando2017bidirectional}, and, \emph{gated recurrent unit} (GRU) \cite{padi2019end}.

Recently, \emph{time delay neural network} (TDNN) based architectures are successfully applied for language recognition \cite{garcia2016stacked}, \cite{miao2019new}, \cite{mandava2019attention}. TDNN models use dilated convolution layers that incorporate the contexts of multiple input frames together. The convolution layers are followed by a pooling layer that produces a fixed dimensional utterance level representation. Usually, some fully connected layers are placed after the pooling layer. Using TDNN, x-vector based embedding was used for language recognition in \cite{snyder2018spoken}. Similar to i-vector, the efficiency of x-vector embedding can further be improved by various post-processing methods. The TDNN architectures are directly used for end-to-end classification as well~\cite{villalba2018end}. However, it was shown in \cite{snyder2018spoken} that using x-vector as a feature extractor and using a Gaussian backend classifier yields better LID performance compared to the end-to-end approach. The language recognition performance with x-vector embedding can be significantly improved by using data augmentation techniques \cite{snyder2018spoken}. The performance of x-vector based systems for short utterances was improved in \cite{shen2020compensation}. Various modified versions of the conventional TDNN models, such as \emph{Factorized TDNN} \cite{povey2018semi}, \emph{Extended TDNN}~\cite{snyder2019jhu}, \emph{ECAPA TDNN}~\cite{desplanques2020ecapa} are also being investigated for the language recognition task. \emph{Attention mechanism} is also used in the pooling layer of the ECAPA TDNN architecture that learns to weight different input frames as per their language discriminating cues. For the sequence models, recently \emph{transformer} architectures are outperforming the RNN and LSTM based models for speech processing applications~\cite{karita2019comparative}. Residual networks,  allowing better gradient flow for longer neural networks, became popular for LID tasks with sufficiently large corpus~\cite{vuddagiri2018improved,shen2020compensation,shen2020knowledge}. In Fig.~\ref{fig:timeline}, we summarize the timeline of various major LID research attempts using different modeling techniques discussed above.  

\subsection{Performance metric}
\label{subsec:metric}
 The performance evaluation for LID systems were analysed in terms of classification accuracy for the earlier systems. The classification accuracy (in $\%$) is computed as:
 \begin{equation}
     accuracy=(\frac{n_c}{n_{tot}})*100
 \end{equation}
 here, $n_{tot}$ denotes the total number of evaluation utterances and $n_c$ indicates the total correctly predicted utterances by a classifier. Accuracy is relatively easy to perceive metric to assess the performance. However, in case of highly class-imbalanced data, accuracy measurement becomes unreliable~\cite{bishop2006pattern}. To deal with this issue, alternative performance metrics; precision, recall, and F1-score are also used in LID. These metrics are usually accompanied with confusion matrix. In Table~\ref{tab:cm}, we have shown the structure for confusion matrix representation.
\begin{scriptsize}
\begin{table}
\vspace{-.02cm}
\centering
\caption{Representation of confusion matrix for a two-class classification problem.}
\vspace{-.15cm}
\label{tab:cm}
\resizebox{.5\linewidth}{!}{%
\begin{tabular}{|l|c|c|} 
\hline
\multicolumn{1}{|c|}{\diagbox{\begin{tabular}[c]{@{}l@{}}Actual~\\class\end{tabular}}{\begin{tabular}[c]{@{}l@{}}Predicted\\class\end{tabular}}} & Negative  & Positive \\ 
\hline
Negative & \begin{tabular}[c]{@{}c@{}}True negative\\(TN) \end{tabular} & False positive~(FP) \\ 
\hline
Positive & \begin{tabular}[c]{@{}c@{}}False negative\\(FN) \end{tabular} & True positive~(TP) \\
\hline
\end{tabular}
}
\vspace{-.45cm}
\end{table}

\end{scriptsize}

Based on Table~\ref{tab:cm}, the alternate metrics are computed:
\begin{equation}
    precision=\frac{TP}{TP+FP}
\end{equation}
\begin{equation}
    recall=\frac{TP}{TP+FN}
\end{equation}
\begin{equation}
    F1=2*\frac{precision * recall}{precision + recall}
\end{equation}
Precision~\cite{sokolova2009systematic} is the useful performance metric when the cost of falsely predicting a class as positive is too high. Similarly, recall~\cite{sokolova2009systematic} is vital for the scenarios where the cost of false negatives are high. F1 is a balanced metric for precision and recall. It is also robust against class-imbalance issues~\cite{haixiang2017learning}.

However, the cost of a classifier is mainly determined by the wrong decisions taken due to both false positives and false negatives~\cite{brummer2006application}. Based upon the detection threshold, the two cost factors create a trade-off scenario. Due to this, in LID literature, cost function based metrics are commonly used which measures the model's effectiveness in minimizing the total weighted cost. LID challenges, such as, \emph{NIST LRE}~\cite{sadjadi20182017,greenberg20122011,martin20102009,martin2003nist}, \emph{OLR challenge}~\cite{li2020ap20,tang2019ap19,tang2017ap17} introduced other performance metrics which became the standard evaluation parameter in this field. \emph{Equal error rate} (EER) and $C_{\mathit{avg}}$ are the most widely used metric. For EER, the \emph{false-acceptance rate} (FAR) and \emph{false-rejection rate (FRR)} values are varied by changing the detection threshold. EER is the minimum value at which the FAR and FRR intersect. The plot with the varying FAR and FRR is known as the \emph{detection error tradeoff} (DET). $C_{\mathit{avg}}$ is defined in Eq.~\ref{eq:cavg} as follows \cite{sadjadi20182017}:
\begin{equation} C_{\mathit{avg}}=\frac{1}{N}\sum\limits_{L_{t}}\begin{Bmatrix} P_{Target}\cdot P_{Miss}(L_{t})\\ +\sum\limits_{L_{n}}P_{Non-Target}\cdot P_{FA}(L_{t}, L_{n}) \end{Bmatrix} 
\label{eq:cavg}
\end{equation}
where, $L_t$ and $L_n$ are the target and non-target languages. $P_{Miss}$, $P_{FA}$ are the probability of missing (FRR) and false alarm (FAR). $P_{Target}$ is the prior-probability of the target languages, usually considered as 0.5. $P_{Non-Target}=(1-P_{Target})/(N-1)$, where, $N$ is the total number of languages. The lower value of EER and $C_{\mathit{avg}}$ indicates better classification performance. 

Both EER and $C_{avg}$ considers a global threshold where the total costs due to the false positives and false negatives are supposed to be optimum. During evaluation, it can be possible that the test utterances have several variations in non-lingual aspects from the training data. Hence for the trained classifier, test utterances will exhibit varying degree of similarity and as consequence, some of them are easier to recognize whereas some are very challenging. Under such circumstances, the prediction scores can largely suffer due to calibration sensitivity. An additional \emph{calibration} stage might be useful here~\citestyle{ferri2009experimental}. Selecting a single threshold value for the calibration-sensitive scores may not yield the optimal operating point by minimizing the total cost of misclassification. Subsequently, both EER and $C_{avg}$ is not calibration-insensitive due to the selection of the single global threshold~\cite{brummer2010measuring}. Further, these metrics were mainly developed for binary classifications. In LID task, we adapt them in multi-class scenario by pooling the individual score component for different target classes. Selection of a single global threshold for different target class scores further increases the calibration dependencies for these metrics~\cite{brummer2010measuring}. These cost-based metrics are often criticized for being over-optimistic; they only consider the prediction scores values and ignore the cost of falsely  predicting the output class, which is very crucial from application perspective~\cite{brummer2006application}. The selection of optimal performance metrics for the LID task is a very promising domain for further investigation.

\vspace{-.1cm}
\section{Overview of Indian Language Recognition \& Challenges}
\label{sec:3}
\vspace{-.05cm}
\subsection{Brief description of languages and linguistic families of India}\label{Sec:3.1}
India has the world's fourth-highest number of languages, after Nigeria, Indonesia, and Papua New Guinea. The Indian languages are further spoken in different kinds of dialects. In the Eighth Schedule of the Indian constitution, 22 widely spoken languages are officially recognized as Scheduled languages: Assamese, Bodo, Dogri, Gujarati, Hindi, Kannada, Kashmiri, Konkani, Maithili, Malayalam, Manipuri, Marathi, Nepali, Odia, Punjabi, Sanskrit, Santali, Sindhi, Tamil, Telugu, and Urdu. 
\begin{figure*}[!t]
\vspace{0cm}
    \centering
    \includegraphics[trim={0cm, 0cm, 0cm, 0cm},clip,width=\textwidth]{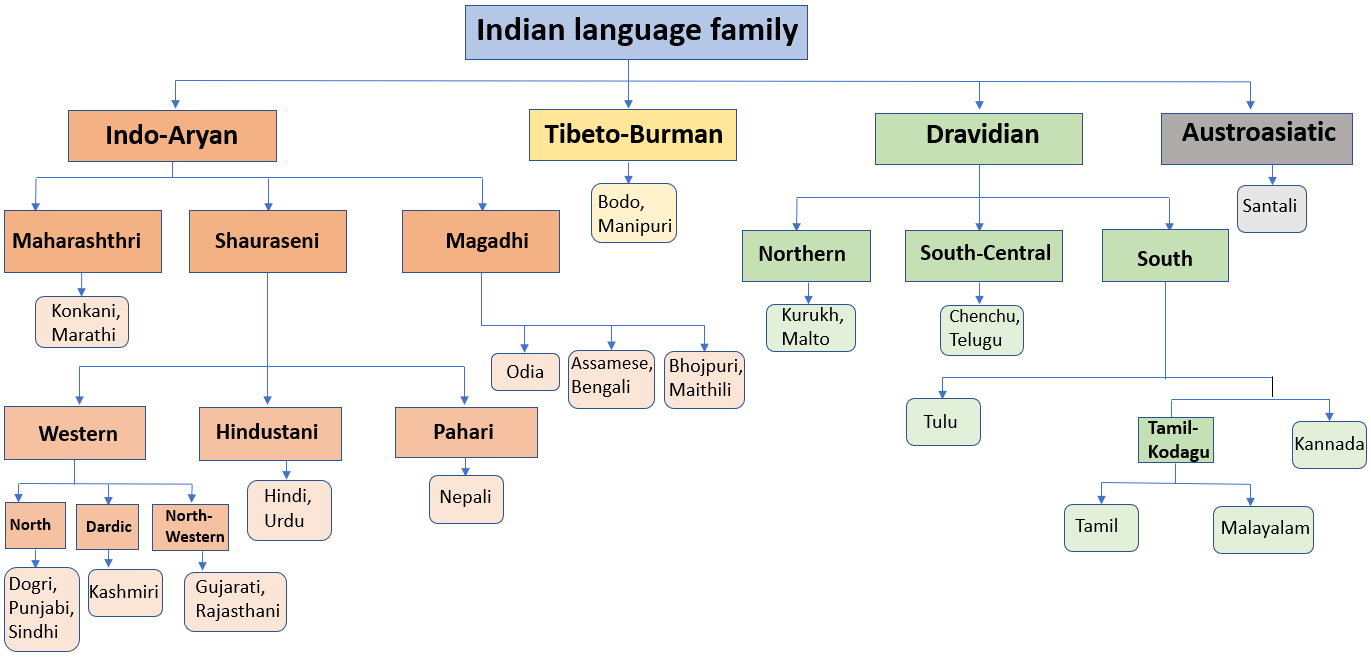}
    \vspace{-.5cm}
    \caption{Detailed linguistic family tree structure for the major Indian languages.}
    \label{fig:lang_family}
    \vspace{-.3cm}
\end{figure*}

As per the linguistic evaluation, Indian languages are mainly classified into the Indo-Aryan family (spoken by $75.05\%$ of the population) and Dravidian family (spoken by $19.64\%$ of the population). The rest $2.31\%$ population belongs to Austroasiatic, Tibeto-Burman, Tai-Kadai, and a few other language families \footnote{\url{https://censusindia.gov.in/2011Census/Language_MTs.html}}. Languages within a language family have evolved from a common ancestral language and possess stronger mutual similarities~\cite{rowe2018concise}. The different language families for Indian languages are summarized in Fig.~\ref{fig:lang_family}. The Indo-Aryan languages have evolved from the Indo-Iranian branch of the Indo-European language family. The Dravidian language family consists of more than 80 languages and dialects, with the major speakers in Telugu, Tamil, Kannada, and Malayalam languages~\cite{kolipakam2018bayesian}. Following~\cite{kamil1990dravidian}, the languages of the Dravidian family can be sub-grouped into South (eg. Tamil, Malayalam, Kannada), South-Central (eg. Telugu), and Northern Dravidian languages (eg. Kurukh, Malto, Brahui). The Tibeto-Burman languages are the \emph{non-Sinitic} branch of the Sino-Tibetan language family. The Austro-Asiatic languages are linguistically divided into two branches; the Mon-Khmer languages, mainly from South-East Asia, and the Munda languages (Santanli, Korku, Mundari, Ho, etc.) from the Central and East regions of India. Aarti et al. in~\cite{aarti2018spoken} provided an organized review about the different linguistic cues according to the major language families in India. Further, in the literature, hierarchical LID systems with a front-end language family recognizer are proposed for improved LID performance~\cite{jothilakshmi2012hierarchical, sengupta2016identification, bakshi2021improving}.
\subsection{Challenges of low-resource}
The term ``low-resourced" or ``under-resourced" languages refers to the languages with lack in some (if not all) factors: lack of writing systems or stable orthography, limited presence in the web, lack of linguistic expertise, lack of electronic resources for developing speech applications~\cite{besacier2014automatic}. A language spoken by a lesser population may not be low-resourced, or a language spoken by millions of speakers can still be low-resourced. Due to the developments of many standard multilingual speech corpora, significant progress has already taken place for language recognition systems in certain languages, such as English, Japanese, Mandarin, Russian, Arabic, etc. The current research trend of applying various state-of-the-art neural network architectures for language recognition also requires a huge amount of multilingual speech database~\cite{gonzalez2015frame}. Due to various spoken language recognition challenges, such as NIST LRE~\cite{martin2003nist,martin20102009,greenberg20122011,martin2014nist,sadjadi20182017}, AP-OLR \cite{tang2017ap17,tang2019ap19,li2020ap20}, etc., a large amount of standard speech corpora was made available to the LID research community. These corpora covered a minimal number of Indian languages. In Table~\ref{tab:nist}, we have shown the Indian languages covered by the major language recognition challenges. Mostly the languages which are spoken in India, as well as other South Asian countries, are provided. Therefore, maximum Indian languages are still \emph{low-resourced} for developing state-of-the-art LID systems. Due to lack of data, LID systems trained with large nerual networks can have poor generalization for real-world deployment. However, in the last few years, attempts of large-scale corpus development are being made for the Indian LID research (discussed in Section~\ref{sec:4}). 
\begin{scriptsize}
\begin{table}
\vspace{-.03cm}
\centering
\caption{Total languages vs. the number of Indian languages covered in the different NIST LRE challenges.}
\vspace{-.03cm}
\label{tab:nist}
\resizebox{\linewidth}{!}{%
\begin{tabular}{|l|c|c|c|} 
\hline
\textbf{Challenge~} & \textbf{Total languages} & \textbf{Indian languages~} & \textbf{Remarks}                      \\ 
\hline
NIST LRE 03         & 14                                 & 2                                    & Hindi and Tamil                       \\
NIST LRE 05         & 7                                  & 2                                    & Hindi and Tamil                       \\
NIST LRE 07         & 13                                 & 4                                    & Bengali, Hindi, Tamil, Urdu           \\
NIST LRE 11         & 24                                 & 5                                    & Bengali, Hindi, Punjabi, Tamil, Urdu  \\
\hline
\end{tabular}
}
\vspace{-.6cm}
\end{table}
\end{scriptsize}

\subsection{Challenges of mutual influence and similarity}
The majority of the Indian languages belong to two language families~\cite{grierson1906linguistic}. The languages within a linguistic family have evolved from the same parental language~\cite{rowe2018concise}. As a consequence, the languages within a family have significant similarities from both linguistic and perceptual perspectives. For example, for Indo-Aryan languages, Bengali and Odia are very similar and mutually intelligible up to a great extent. In our previous work~\cite{sengupta2016identification}, we showed that due to geographical proximity and cultural interactions, even the languages from different language families have a noticeable influence on each other. Almost all of the Indian languages are non-tonal, syllable-timed \cite{emeneau1956india}. The mutual influence and similarity among the languages have made Indian spoken language recognition challenging. Sophisticated approaches, either in signal level, feature level, or classifier level, are needed to discriminate among easily confusing languages. For example, most Indian languages are non-tonal, whereas Punjabi and some North-Eastern Indian languages show tonal characteristics. Along with the conventional acoustic features, if prosodic features are incorporated into the LID system, the language discrimination capability increases \cite{bhanja2019pre}. 

\begin{figure}[t]
    \centering
    \includegraphics[trim={0cm .1cm .3cm 0cm},clip,width=.9\textwidth]{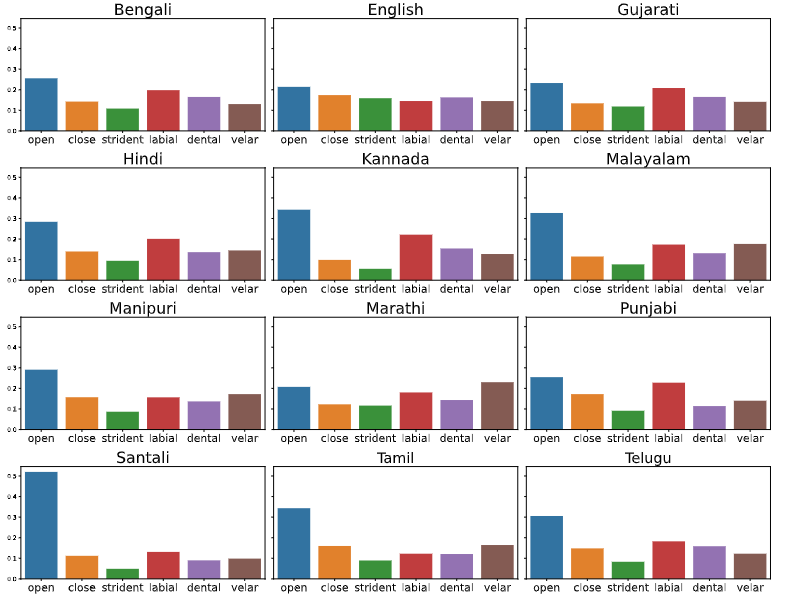}
    \vspace{-.05cm}
    \caption{Comparison for the gross likelihoods of the six phoneme sub-classes as per the occurrences in 12 Indian languages from IIITH corpus.}
    \label{fig:phone_histo}
    \vspace{-.75cm}
\end{figure}

From phonological perspectives, apart from Indian English and Urdu, the rest of the Indian languages share the same phoneme repository \cite{maity2012iitkgp}. Although, the pronunciation of the same phoneme may be slightly different across the languages, even with common phoneme repositories. For that reason, we have attempted to plot the phoneme occurrence likelihoods of several Indian languages by categorizing the phonemes into several sub-classes. We have used 12 Indian languages from the IIITH-ILSC~\cite{vuddagiri2018iiith} corpus. The phonemes are categorized into six sub-classes according to the manner and place of articulation using~\cite{vasquez2019phonet}. We compute the likelihood of the six sub-classes for non-overlapping speech segments of 20~ms and then average it over the entire utterance. Finally, we then compute the gross phoneme sub-class likelihoods for each language by averaging across the utterances. The language-wise gross likelihoods are normalized and plotted in Fig.~\ref{fig:phone_histo}. The patterns of the likelihoods for all the languages are almost the same except for Indian English and Santali. Even though Indian English has a different accent from the UK or US English, but it is not originated in India. Therefore, the patterns of the phoneme likelihoods can be different. However, as also observed in Fig.~\ref{fig:phone_histo}, Indian English is not drastically different from other Indian languages due to the L1 influence~\cite{wiltshire2006influence,vijayarajsolomon2017exploiting,maxwell2010acoustic}. Santali, belonging from the Austroasiatic language family, differs in phoneme likelihood pattern from the other languages of Indo-Aryan and Dravidian families. Except for these two cases, for all the other languages, the phone-class likelihoods are very similar, indicating the close linguistic similarity among the Indian languages.

\section{Speech Corpora for Indian Language Recognition}
\label{sec:4}
In this section, we first describe the desired characteristics of the corpora for spoken language recognition research. Then, we briefly review the most widely used corpora for the LID research in the context of the Indian languages. 
\subsection{Characteristics of standard speech corpora for LID research}
The collection of audio data for major Indian languages is not very challenging as before due to the availability of the Internet. However, for the LID evaluation, any audio data containing Indian languages may not be appropriate. Speech corpora for language recognition tasks should follow certain criteria:
\begin{itemize}
    \item The speech data should be recorded in environments with varying backgrounds. It ensures robustness against varying background noises for the models trained with the corpus. ~\cite{hansen2018issues}.
    \item To avoid speaker-dependent bias, for each language, data from a large number of speakers should be collected~\cite{hansen2018issues}.
    \item The number of male and female speakers should be balanced. This requirement is essential for unbiased LID performance across the genders.
    \item Ideally, there should not be significant bias for the acoustic room environments among the utterances of different language classes. If this criterion is not fulfilled, the classifier model can recognize different recording environments as different language classes~\cite{sturm2014simple}.
    \item The variations due to several transmission channels should also be taken care of such that these variations should not be confused with individual language identities.
    \item The speakers for each language should ideally cover different age groups~\cite{behravan2015factors}. 
    \item In order to incorporate the dialect and accent variations \cite{biadsy2011automatic}, for each language, speakers from different geographical areas and social-cultural backgrounds should be taken into consideration~\cite{hansen2018issues}.
    \item The most common sources of speech data in the developed corpora are broadcast news (BN), interviews, recorded TV programs. In these sources, the speakers generally use the standard form of the spoken language. The manner of articulation is also very professional. Whereas in spontaneous conversations, for example, \emph{conversation telephone speech} (CTS), there may be significant dialect and accented variations. The manner of articulation is not restricted to being professional. The emotional variations are also frequent in CTS sources. The desired corpora should collect speech from both the broadcast and conversation sources~\cite{gonzalez2010multilevel}.
\end{itemize}
\subsection{Review of major corpora available for Indian LID}
Speech corpora consisting of Indian languages have been developed for several purposes, such as speech recognition, speaker recognition, speech synthesis, translation, etc. There are several organizations, such as \emph{Central Institute of Indian Languages} (CIIL, Mysore, India)~\footnote{\url{https://www.ciil.org/}}, \emph{Center for Development of Advanced Computing} (C-DAC, India)~\footnote{\url{https://www.cdac.in/}}, \emph{The Linguistic Data Consortium for Indian Languages} (LDC-IL)~\footnote{\url{https://ldcil.org/}}, along with the educational institutes, that are actively involved in creating several standard multilingual Indian speech corpora. The language recognition challenges, such as NIST LRE~\cite{martin2003nist,martin20102009,greenberg20122011,martin2014nist,sadjadi20182017} and AP-OLR~\cite{tang2017ap17,tang2019ap19,li2020ap20}, have also contributed in providing speech data in some of the most widely spoken languages of India and South Asia. Here we briefly describe the most widely used corpora for the Indian LID research.
\subsubsection{Initial developments:}
The \emph{EMILLE (enabling minority language engineering) / CIIL} Corpus \cite{xiao2004developing}, is one of the initial attempts to develop standard speech corpora that can be used for Indian LID task. This database contains three sections, monolingual, parallel, and annotated data. The monolingual speech data contains more than 2.6 million words in several Indian languages, such as Bengali (442k words), Hindi (588k words), Gujarati (564k words), Punjabi (521k words), Urdu (512k words). Another corpus was developed by C-DAC Kolkata, which consisted of annotated speech data in three eastern and North-Eastern Indian languages, Bengali, Assamese, and Manipuri. For the spoken utterances, syllables and breath pauses have been annotated. The data were recorded by professional artists, and only the standard dialect of a particular language is used. The \emph{OGLI-MLTS} corpora~\cite{muthusamy1992ogi} contained 1545 telephone conversations in 11 languages. Two Indian languages, Hindi (included in updated version) and Tamil, were provided among these languages. For Hindi, 200 calls and Tamil 150 calls are included, with an average call duration of approximately 80 seconds. In the NIST LRE 03 and 05, Hindi and Tamil corpora were included. Later on, in LRE 11, five Indian languages, Bengali, Hindi, Punjabi, Tamil, and Urdu, were included. These are some of the earlier available resources for Indian LID research. These databases had several limitations:
\begin{itemize}
    \item The number of Indian languages available for LID research was minimal.
    \item Even for those languages, the amount of speech data provided was not extensively vast.
    \item Majority of the developed corpus was for other speech-based applications, such as speech recognition. Very few attempts were made to develop dedicated speech corpora for Indian LID research.
    \item In many cases, some fixed keywords or a limited number of sentences were used as utterances. It could make the developed LID systems more context-dependent.
\end{itemize}
\subsubsection{IITKGP-MLILSC:}
This corpus was developed and presented in~\cite{maity2012iitkgp} to deal with some of the above-mentioned drawbacks. This was the first corpora that covered a larger number of Indian languages. This corpus contained speech data of total 27 hours in 27 Indian languages: Arunachali (72 minutes), Assamese (67.33 minutes), Bengali (69.78 minutes), Bhojpuri (59.82 minutes), Chattisgarhi (70 minutes), Dogri (70 minutes), Gojri (44 minutes), Gujarati (48.96 minutes), Hindi (134.70 minutes), Indian English (81.66 minutes),
Kannada (69.33 minutes), Kashmiri (59.64 minutes), Malayalam (81.09 minutes), Marathi (74.33 minutes), Nagamese (60 minutes), Nepali (54.19 minutes), Oriya (59.87 minutes), Punjabi (80.91 minutes), Rajasthani (60 minutes), Sanskrit (70 minutes), Sindhi (50 minutes), Tamil (70.96 minutes), Telugu (73.72 minutes), and Urdu (86.49 minutes). The audios were recorded from TV and radio broadcasts. Non-speech distortions, such as background music during news headlines and advertisement breaks, overlapping speech, were manually discarded. The development of this corpus was very important for LID research in many of the low-resourced Indian languages.
\subsubsection{IIITH-ILSC:}
In a similar spirit, recently, another multilingual speech corpus \emph{IIITH-ILSC} \cite{vuddagiri2018iiith},  consisting of 23 Indian languages, was introduced. It contains all the 22 official languages of Indian along with Indian English. The recent research trend encourages to use neural network architecture for LID purposes. However, for efficiently implementing neural networks, a larger amount of speech data is required for the language classes~\cite{snyder2018spoken}. From this perspective, apart from covering many Indian languages, the IIITH-ILSC database covered a significantly larger amount of data per language compared to the earlier proposed speech databases for Indian LID. This database had a total of 103.5 hours of speech data. For each language, 4.5 hours of data was present, 3.5 hours for training-validation and 1 hour for testing. Each language contained data from 50 speakers, including 25 males and 25 females. Both read speech and conversational data were present, and the audio quality varies from clean speech to moderately noisy speech. Presently this is one of the largest speech corpora used for Indian spoken language recognition research. 
\subsubsection{LDC South Asian corpus (LDC2017S14):}
LDC has provided 118 hours of conversational telephone speech corpus~\cite{LDC} with five widely spoken languages in the Indian sub-continent: Bengali (26.6 hours), Hindi (7.4 hours), Punjabi (western) (38.8 hours), Tamil (22.9 hours), and Urdu (22.9 hours). The total duration of this database was almost 118 hours. Part of this database was used in the NIST LRE 2011 challenge \cite{greenberg20122011}. The amount of data is suitable enough for neural network based architectures. While most Indian multilingual databases contain broadcast speech, this corpora provided spoken language data from spontaneous conversation by native speakers.
\subsubsection{VoxLingua107:}
Very recently, similar to the \emph{VoxCeleb} corpora \cite{nagrani2020voxceleb} in speaker recognition, a huge speech corpora \emph{VoxLingua107} is prepared for LID purpose \cite{valk2021slt}. This corpus contains 6628 hours of speech data in 107 languages, collected from YouTube. This corpora also contains several Indian languages, such as Assamese (155 hours), Bengali (55 hours), Gujarati (46 hours), Hindi (81 hours), Kannada (46 hours), Malayalam (47 hours), Marathi (85 hours), Nepali (72 hours), Punjabi (54 hours), Sanskrit (15 hours), Sindhi (84 hours), Tamil (51 hours), Telugu (77 hours), Urdu (42 hours). This corpus can be very promising for future research in developing Indian LID systems with complex state-of-the-art classifier architectures.
\subsubsection{Other developments:}
There are some other speech corpora developed in the last few years that contain Indian languages. The \emph{Microsoft speech corpus for Indian languages} contained speech data in Gujarati, Tamil, and Telugu. \emph{NITS-LD} speech corpora used in \cite{bhanja2019deep}, contained speech data collected from All India Radio broadcasts in seven Indian languages, Assamese, Bengali, Hindi, Indian English, Manipuri, Mizo, and Nagamese. This database collected almost four hours of speech per language. Recently in~\cite{basu2021multilingual}, a speech corpus with 16 low-resourced North-East Indian languages is presented. This corpus contains speech data of some rarely studied Indian languages, such as Angami, Bodo, Khasi, Hrangkhawl, and Sumi. Some of the other sources for availing standard Indian speech data are \emph{Speehocean}~\footnote{\url{ http://www.speechocean.com}}, \emph{Indic-TTS}~\footnote{\url{" https://www.iitm.ac.in/donlab/tts/index.php}}~\footnote{"Indic" languages are referred to the Indo-Aryan languages. Using this term to denote all the Indian languages is a misnomer.}. There also developments in \emph{open-source corpora}, such as \emph{Mozilla Common Voice}~\footnote{\url{https://commonvoice.mozilla.org/en}}, \emph{OpenSLR}~\footnote{\url{https://www.openslr.org/}} with speech data for the Indian languages. In Table~\ref{tab:database_compare}, we have summarized the key information about the major speech corpora developed for Indian spoken language recognition research.

\section{Literature Review of Indian Spoken Language Recognition} 
\label{sec:5}
This section presents a brief overview of the major research works in developing Indian spoken language recognition systems. The fundamentals of Indian LID research follow the primary trends of LID research in a general sense. However, considering the unique challenges for Indian languages, researchers adapt the feature extraction and classification stages accordingly. Apart from language classification, there are studies on classifying language families, detecting language similarities and mutual influence. There are few prior research attempts that surveyed the database and language-dependent features for the Indian LID~\cite{aarti2018spoken,shrishrimal2012indian}. The survey of the proposed methodology for the different Indian LID systems were not focused. Contrary to the prior review attempts, in this work we aim to conduct an in-depth system development analysis from the machine learning perspectives. In the beginning, the earlier research attempts are briefly discussed. Analyzing their limitations, we then provided an in-depth description and critical analysis of the relatively recent research works. As most research works considered different corpora, different languages, different data split and cross-validation protocols, and various performance metrics, we sincerely avoid numerical comparisons of the reported numbers.
\begin{table}[!htbp]
\centering
\caption{Summarized description of some widely used corpora for Indian LID.}
\label{tab:database_compare}
\resizebox{.8\linewidth}{!}{%
\begin{tabular}{|>{\hspace{0pt}}m{0.1\linewidth}|>{\centering\hspace{0pt}}m{0.11\linewidth}|>{\centering\hspace{0pt}}m{0.12\linewidth}|>{\centering\hspace{0pt}}m{0.15\linewidth}|>{\centering\hspace{0pt}}m{0.152\linewidth}|>{\centering\hspace{0pt}}m{0.095\linewidth}|>{\hspace{0pt}}m{0.094\linewidth}|} 
\hline
\textbf{Corpora} & \textbf{IITKGP-MLILSC~\cite{maity2012iitkgp}} & \textbf{LDC South Asian~\cite{LDC}}\par{}\textbf{(From NIST LRE 11)} & \textbf{IIITH-ILSC~\cite{vuddagiri2018iiith}} & \textbf{VoxLingua107~\cite{valk2021slt} (Indian languages only)} & \textbf{NITS-LD~\cite{bhanja2019deep}} & \textbf{EMILLE /CIIL~\cite{xiao2004developing}} \\ 
\hline \hline
\textbf{No. of languages} & 27 & 5 & 23 & 14 & 12 & 5 \\ 
\hline
\textbf{Avg speakers/language} & 10 & 110 & 50 & NA & 18 & NA  \\ 
\hline
\textbf{Mode of speech} & Broadcast & CTS & Broadcast and CTS & Automatically collected web audio data & Broadcast news & Monolingual words \\ 
\hline
\textbf{Environment} & Studio, room, outdoor  & Room, outdoor & Studio, room & Collected from wild sources & Studio  & Room \\ 
\hline
\textbf{Noise} & Low background\par{} noise & Both clean and\par{} moderately noisy & Both clean and\par{} moderately noisy & Diverse noise level & Less noisy & Less noisy \\ 
\hline
\textbf{Channel variation} & Less & Exists & Exists &  Exists &  Less & Less \\ 
\hline
\textbf{Avg Hours / language} & 1 hour & $\sim$22 hours & 4.5 hours & 65 hours & 11.5 hours & NA \\ 
\hline
\textbf{Total no. of speakers} & 300 & 584 & 1150 & NA & 212 &  NA  \\ 
\hline
\textbf{Total Hrs of speech} & 27 hours & 118.3 hours & 103.5 hours & 910 hours & 136 hours & NA \\ 
\hline
\textbf{Audio format} & 8 kHz (.wav) & 8 kHz (.flac) & 16 kHz (.wav) & 16 kHz (.wav) & 8 kHz  & 8 kHz \\
\hline
\textbf{Data-split} & Train-test (80:20) & Not provided & Train-validation-test (70:10:20) & Not provided & -  & - \\

\hline
\end{tabular}
}
\end{table}

\subsection{Brief overview of earlier research attempts for Indian LID}
Jyotsna et al. (2000)~\cite{balleda2000language} made one of the earliest attempts to classify spoken Indian languages automatically. Five Indian languages, Tamil, Telugu, Kannada, Malayalam, and Hindi, were used. Vector quantization method was used to build the language-specific classifier model. 
In Kumar et al. (2004)~\cite{kumar2004language}, LID experiment was performed for two Indian languages, Hindi and Tamil. The authors explored phonotactic characteristics of the languages for classification.
Mary et al. (2004)~\cite{mary2004autoassociative} used an \emph{auto-associative neural network} (AANN) based model for performing LID task using four Indian languages, Hindi, Tamil, Telugu, and Kannada. \emph{weighted linear prediction cepstral coefficient} (WLPCC) feature was used to train the AANN model. Manwani et al. (2007)~\cite{manwani2007spoken} used GMM based modeling technique for performing LID with Hindi, Telugu, Gujarati, and English. MFCC with delta ($\Delta$) and acceleration ($\Delta^2$) was used. Instead of using the conventional expectation-maximization (EM) based method, the authors used the split and merge EM algorithm for GMM training.  
The classification accuracy was improved as compared to EM-based GMM models. Mohanty et al. (2011)~\cite{mohanty2011phonotactic} used parallel phone recognizer (PPR) based language modeling followed by SVM for language classification. Four Indian languages, Bengali, Hindi, Odia, and Telugu, were used for the LID task. 
Jothilakshmi et al. (2012)~\cite{jothilakshmi2012hierarchical} developed a two-stage Indian LID system. The first stage was for identifying the language family, and then, in the second stage, separate language recognizer models were trained for the languages from each family. MFCC feature was used with SDC and delta coefficients. Several classifier techniques, HMM, GMM, ANN, were used for comparison.

After thoroughly inspecting the earlier research attempts for developing Indian LID systems, we figured out major challenging issues for conducting research in the Indian LID domain. These are summarized in Table~\ref{tab5}. Some of them are successfully addressed in the recent research works, and some issues still exist and could be a potential topic for future research direction.

\begin{scriptsize}
\begin{table}
\vspace{-.1cm}
\centering
\caption{Summary of the issues in Indian LID task.}
\vspace{-.05cm}
\label{tab5}
\resizebox{\linewidth}{!}{%
\begin{tabular}{|l|l|} 
\hline
\textbf{Challenge type} & \textbf{Description} \\ 
\hline
\textbf{Data} & \begin{tabular}[c]{@{}l@{}}1. Lack of significant corpora development activities.\\2. Limited availability or scarcity of speech data for some languages.\\3. Limited non-lingual variations in the data resulting poor generalization.\end{tabular} \\ 
\hline
\textbf{Architectural} & \begin{tabular}[c]{@{}l@{}}1. Computational limitation was a major constraint in the earlier research attempts. \\2. Lacks exploration of large efficient architectures due to limited training data.\\3. Promising end-to-end architectures is not explored widely.\end{tabular} \\ 
\hline
\textbf{Application} & \begin{tabular}[c]{@{}l@{}}1. Performance reduction real-world applications with short test utterances.\\2. Performance degradation in discriminating highly confusing Indian languages.\\3. Adaptation for dialect, accent variations, and code-switching effects.\\4. Overlooking the importance of the open-set evaluation conditions.\end{tabular} \\
\hline
\textbf{Experimental} & \begin{tabular}[c]{@{}l@{}}1. Lack of adoption of fixed evaluation metrics.\\2. Lack of fixed data split and standard evaluation protocols.\end{tabular} \\
\hline

\end{tabular}
}
\vspace{-.5cm}
\end{table}
\end{scriptsize}


\subsection{Literature review of relatively recent research works for Indian LID}
Through the advancement of time, several standard speech corpora have been developed that cover a large number of major Indian languages. Alongside, the amount of speech data available for each of the languages have also increased dramatically. The machines' computation power has also improved, primarily due to graphical processing units (GPUs). From VQ, Phone-based language modeling, followed by GMM and SVM, the paradigm of machine learning modeling has gradually shifted towards i-vector, and most recently, neural network based architectures. Considering all these facts, we have presented an independent, in-depth analytical review of the relatively recent research attempts for Indian LID.

Maity et al. (2012)~\cite{maity2012iitkgp} introduced the IITKGP-MLILSC speech database and performed the LID task using both speaker-dependent and independent manners. Earlier, the research of Indian spoken language recognition was limited to 4-5 major languages due to limitations of available speech resources. This corpus had presented speech utterances from 27 Indian languages. Out of which, the authors used 16 most widely spoken Indian languages for the LID task. GMM with mixture coefficients 32 to 512 was used for the modeling technique. 13-dimensional MFCC and 13-dimensional LPCC features were used to build separate language models, and their LID performance was compared. LID task was performed in two manners: in speaker-dependent approach, language data from all the speakers was used for training and testing. Whereas in a speaker-independent way, the speakers used in testing were not used in training data. Test utterances of duration 5, 10, and 20 seconds were used. It was found that LID performance improved as the number of GMM mixture coefficients are increased from 32 to 256. Beyond that, the performance improvement was negligible. MFCC performed better than LPCC features. Although, for smaller GMM mixtures, LPCC performed better.
The speaker-dependent LID model had significantly outperformed the speaker-independent models because some of the speaker information of the test utterances are already known during training. However, when the speakers are entirely arbitrary in a real-time scenario, the speaker-dependent model may not generalize well.

Verma et al. (2013)~\cite{verma2013indian} performed LID on Hindi, English, and Tibetan using MFCC feature and SVM classifier. For each language, 3 minutes of audio recording was collected from 11 speakers. 24-dimensional MFCC was extracted using 30 milliseconds (ms) of window and $50\%$ overlap. This configuration resulted in 4000 MFCC samples for 1 minute of speech data. To reduce the complexity of the LID system, the authors reduced the number of MFCC samples using K-means clustering. It resulted in reduced training time with nominal performance degradation.

In ~\cite{reddy2013identification}, Reddy et al. (2013) experimented with three approaches for extracting MFCC, block processing (BP), pitch synchronous analysis (PSA), and glottal closure instants (GCI). In BP, the speech segment was framed using 20 ms of window with an overlap of 10 ms between consecutive frames. In PSA, MFCC was extracted only from individual pitch cycles. One pitch cycle was identified as the segments in between consecutive glottal closure instants (GCI). In GCR, only $30\%$ of the pitch period was taken from both sides of GCI points. They used the IITKGP-MLILSC corpus and applied GMM classifier for performing LID. Further, they extracted prosodic features in syllable, tri-syllable (word), and phrase (multi-word) level. Syllables were detected as the portions between consecutive \emph{vowel onset points} (VOPs). From each syllable, seven-dimensional prosodic features describing intonation, rhythm, and stress were extracted. Prosody for three consecutive syllables was concatenated to make 21-dimensional tri-syllabic features. For phrase level, pitch contour, syllable duration, and energy contour of 15 consecutive syllables were concatenated. The prosody-based LID system was then used for score fusion with the MFCC-based LID system. Prosodic features contained complementary language discriminating information as compared to MFCC. Hence the fused system further improved the LID performance compared to the standalone MFCC systems. The authors reported prosody features improving the noise robustness. GCR-based MFCC feature performed the best, followed by PSA and the conventional BP approach. PSA captured the finer acoustic variations within one pitch cycle as compared to BP. In GCR, only the high SNR regions within a pitch cycle were used.

Bhaskar et al. (2013)~\cite{bhaskar2013analysis} used the IITKGP-MLILSC corpus to study the gender dependency on LID performance. First, they built two separate LID models using the male and female speakers, respectively. During testing, they averaged the scores from the two models. For comparison, they combined the speakers of both genders and built the gender-independent model. GMM was used with MFCC for training the models. They showed that the gender-dependent models performed better. A trained gender recognizer model can also be placed before the gender-dependent LID systems.

Alongside the conventional MFCC features, speech information from other sources was also explored for Indian LID. In Nandi et al. (2015)~\cite{nandi2015implicit} the authors explored the excitation source-related acoustic feature for the LID task using the IITKGP-MLILSC corpus. LP residual signal of order ten was used to represent the excitation source. Along with the raw LP residual, its \emph{Hilbert envelope} (HE) and residual phase (RP) were also used to represent the excitation source information. All the features were extracted from subsegmental, segmental, and suprasegmental speech frames. For comparison, vocal-tract-related MFCC features are also used. GMM was used for language modeling. For an MFCC-based system, with a reduction of test utterance duration, LID performance rapidly degrades. However, the authors reported that the excitation source features are more robust to the test utterance duration. The authors also added different noises of 5 to 30 dB with the test data and reported the excitation features, especially the segmental residual phase was significantly noise-robust for the LID task. The speech production system comprises two components; the excitation source due to vocal-folds vibration followed by the vocal-tract system. During the conventional MFCC extraction, the first component is neglected, which provides complimentary language discriminating information by score-fusion. Veera et al. (2018)~\cite{veera2018combining} also explored the excitation source related features for Indian LID but with DNN and DNN with attention (DNN-WA) architectures. They reported the DNN-WA model outperforming i-vector and DNN-based baseline systems. Dutta et al. (2018)~\cite{dutta2018language} explored several phase spectrum-based features: group delay based cepstral coefficient (GD), auto-regressive model based group delay cepstral coefficient (ARGD), and auto-regressive based group delay with scale factor augmentation (ARGDSF) for LID task using the IITKGP-MLILSC corpus and GMM classifier. The conventional MFCC feature uses the magnitude spectrum after Fourier transform. The phase spectrum information is not included in it. The authors showed that the LID performance with phase information is also in the competitive range of the MFCC features. These alternative acoustic features have shown promising LID performances and can be useful for future research directions. However, the feature extraction procedures for these features are more complex as compared to the MFCC. The standalone LID systems trained with such features are not commonly used, and they are needed to be fused with the conventional LID systems for improved performance.

Sengupta et al. (2015)~\cite{sengupta2015study} built a LID system with self-collected speech data from broadcast news in 16 Indian languages; Assamese, Bengali, English, Gujarati, Hindi, Kannada, Kashmiri, Konkani, Malayalam, Marathi, Odia, Punjabi, Sanskrit,
Tamil, Telugu, and Urdu. They used MFCC and SFCC (speech signal-based frequency cepstral coefficient) with SDC features and for each them trained GMM and SVM-based systems. For testing, utterances of 3, 10, and 30 seconds were used. SFCC based models were reported to have slightly better EER as compared to the MFCC based models. Further, they used the false acceptance rate (FAR) of the trained systems to detect the similarity among the languages. If any target, non-target language pairs have higher FAR, then they suggested those languages to be similar. Finally, they identified some highly similar Indian languages, which are often misclassified and degrade overall LID performance. In \cite{sengupta2016identification}, the same authors developed a language family identification model which classified the same 16 Indian languages into Indo-Aryan and Dravidian families. The authors then attempted to find out the influence of one language family onto the individual languages of the other family. For example, suppose the influence of the Dravidian language family is to be checked on language A, belonging to the Indo-Aryan family. During training, training data is considered from all the 16 Indo-Aryan languages except language A. The training data for the Dravidian family remained the same as earlier. During testing, only utterances from language A were used. If a significant portion of test utterances of language A were misclassified as Dravidian language, then the authors claimed that the other language family significantly influenced language A. These works encourage the application of LID systems for analytical studies and validation of the existing linguistic theories.
 
Mounika et al. (2016)~\cite{mounika2016investigation} collected 150 hours of speech data for 12 Indian languages (Assamese, Bengali, Gujarati, Hindi, Kannada, Malayalam, Manipuri, Marathi, Odia, Punjabi, Telugu, and Urdu.) and utilized this large data in DNN architectures. They used 39-dimensional MFCC-$\Delta-\Delta^2$ features. They experimented with different DNN architectures and showed that four hidden layers achieved the best EER for this task. They further extended this DNN with attention (DNN-WA) which outperformed the DNN model. From the weights trained on the attention network, the relative importance to all the frames for making the utterance level prediction was revealed in this work. The authors were able to show that the frames where speech events are in transition carried higher attention weights.

The frame length for MFCC extraction was optimized based on human-level LID performance in ~\cite{aarti2017spoken} by Aarti et al. (2017). They used 75 hours of self-collected speech data over 9 Indian languages from \emph{All India Radio} news. Instead of the convention 25 ms frame length, they showed that with 100 ms frame length, the MFCCs captured better language discriminating information. In a later work~\cite{bakshi2021gmm}, the authors trained a GMM model with MFCC-$\Delta-\Delta^2$ features and used the GMM supervectors as inputs to train the DNN based LID model. 

In~\cite{madhu2017automatic}, LID task was performed with seven Indian languages, Assamese, Bengali, Hindi, Manipuri, Punjabi, Telugu, Urdu. The authors developed two LID systems based on phonotactic and prosodic features, respectively. Phoneme-sequence of two consecutive syllables is used for phonotactic information extraction. Intonation, rhythm, and stress were used for prosodic information. Both the features were fed to DNN architectures. The phonotactic system achieved better LID performance as compared to the prosodic system.

Vuddagiri et al. (2018)~\cite{vuddagiri2018iiith} introduced the \emph{IIITH-ILSC} corpus for Indian LID research. This corpus contains 103.5 hours of speech data in 23 Indian languages. On average, 4.5 hours of speech data was available for each language. It is one of the largest standard speech corpora developed for Indian LID research. The authors then extracted 39-dimensional MFCC-$\Delta-\Delta^2$ features and built LID models based on i-vector, DNN, and DNN-WA. Their results showed that the amount of data available in this corpus is suitable for training the deep neural networks based architectures.

Bhanja et al. (2019)~\cite{bhanja2019pre} used two-stage LID systems based on seven North-Eastern Indian languages from the NITS-LD corpus. The first stage classified the languages into tonal and non-tonal languages. Two parallel LID blocks were trained in the second stage, one for tonal languages and another for non-tonal languages. The use of the pre-classification stage helps to distinguish closely related languages. Multistage LID is also useful when the number of target languages increases. For the pre-classification stage, along with 35-dimensional MFCC features, 18-dimensional prosody features, based on pith contour, energy contour statistics, and syllable duration was proposed. For the second stage LID task, the proposed prosodic features were used along with 105-dimensional MFCC+$\Delta+\Delta^2$ features. For both the tasks, three modeling techniques, GMM-UBM, i-vector with SVM, and, ANN was used. For the pre-classification stage, ANN, and for the LID stage, GMM-UBM yields the best overall performance. However, the overall LID accuracy in this work depends on the efficiency of the first stage that detects the tonal languages. In a similar approach, the authors also developed another LID system~\cite{bhanja2019deep}, which examined the phase-based \emph{Mel-Hilbert envelope coefficient} (MHEC) features along with MFCC features instead of the prosodic features. They followed the BP, PSA, and GCR extraction approaches and extracted 56-dimensional MFCC+SDC features. They extended it to 280 dimensions by fitting into five coefficients using $4^{th}$ order Legendre polynomials. Similarly, they also extracted 280-dimensional MHEC+SDC features. Similar to the observations in~\cite{reddy2013identification} GCR approach performed the best. The authors further showed that the MHEC features performed better for noisy test utterances and tonal languages. The authors also used cascaded CNN-LSTM architecture with features from pitch chroma and formant frequencies in~\cite{china2020cascade}, which further improved the LID performance. In~\cite{bhanja2021modelling}, they used tonal and non-tonal based hierarchical LID system for the prosody and MHEC features in multi-level analysis; syllable, word, and phase.

Embedding from LSTM-\emph{connectionist temporal classifier} (CTC) based ASR model was used for the LID task by Mandava et al. (2019)~\cite{mandava2019attention}. Along with the ASR bottleneck feature, they used MFCC with SDC and trained DNN based LID models. They used the IIITH-ILSC corpus. In the DNN architecture, they used single-head and multi-head attention techniques. The authors showed that CTC features recognized languages better than the SDC feature. The trained attention-based DNN model also outperformed LSTM based sequence modeling techniques for the Indian LID task. The same authors in ~\cite{mandava2019attention} applied an attention-based residual TDNN architecture for the Indian LID task. They showed that the implemented model outperformed LSTM based models for the LID task.

Das et al. (2020)~\cite{das2020bottleneck}, experimented LID task on Assamese, Bengali, Hindi, and English. They first extracted 13-dimensional MFCC, 13-dimensional SDC, and six-dimensional LPC features. The stacked 35-dimensional input features were used to extract ten-dimensional bottleneck features. The bottleneck feature trained a deep auto-encoder followed by softmax regression using the Jaya optimization algorithm. Paul et al. (2020) used the IIITH corpus and developed LSTM based LID classifiers for seven Indian languages: Bengali, Hindi, Kannada, Malayalam, Marathi, Tamil, and Telugu.

In~\cite{mukherjee2019deep}, Mukherjee et al. conducted LID experiments with IIITH-ILSC corpus. They used spectrogram as an input feature to train a CNN LID model. Mukherjee et al. (2020)~\cite{mukherjee2020lazy} further proposed a modified MFCC feature for developing a LID model with Bengali. Hindi, and English. They collected data from online video streaming sites. The authors applied the modified MFCC features on a lazy learning based classifier which is computationally more efficient than the neural architectures. The modified MFCC feature reduced the number of feature coefficients and provided utterance level predictions. Further, the authors showed that the performance degradation with the reduction in utterance length is also less with the modified MFCC.

Das et al. (2020)~\cite{das2020hybrid} proposed a nature-inspired feature selection approach utilizing the \emph{binary bat algorithm} (BBA) with \emph{late acceptance hill-climbing} (LAHC) algorithm. They used 10 Indian languages from the Indic-TTS corpus. Different combinations of the commonly used features, MFCC, GFCC, LPC, x-vector embedding, were used. The combination of MFCC with LPC performed the best.

In~\cite{siddhartha2020language}, the authors experimented with the sub-band filters for extracting excitation source information for the Indian LID task. They used IIITH-ILSC corpus and trained DNN with attention architecture. They extracted the LP residual and applied Mel sub-band filters (RMFCC) and uniform triangular sub-band filters (RLFCC). The language recognition performance showed that the RLFCC extracted more language discriminating information than the conventional RMFCC.

Deshwal et al. (2020)~\cite{deshwal2020language} performed language recognition on four Indian languages, English, Hindi, Malayalam, and Tamil. They proposed hybrid features by concatenating different combinations of MFCC, PLP, and RASTA-PLP and applied them to train a DNN classifier. The hybrid features showed better LID performance. Among the hybrid feature, the combination of MFCC with RASTA-PLP performed the best.

Bottleneck features were explored for the LID task in four Indian languages, Assamese, Bengali, English, and Hindi, by Das et al.~\cite{das2020bottleneck} (2020). The authors first extracted MFCC and LPC features with SDC and stacked them. Then they trained a DNN architecture with the stacked features and extracted the bottleneck features. Finally, a deep autoencoder network was trained on the bottleneck feature for the LID task. The deep autoencoder was trained to achieve the global optimum using the \emph{Jaya optimization algorithm.}

Garain et al. (2020)~\cite{garain2021fuzzygcp}, implemented DNN, CNN, and semi-supervised generative adversarial network (SS-GAN) for LID tasks and performed ensemble learning with the Choquet fuzzy integral method. They used MFCC and spectral statistics such as bandwidth, contrast, roll-off, flatness, centroid, tone and extended them with respective polynomial features. Two Indian databases from IIITH and IIT Madras were used, along with two global databases, VoxForge and MaSS were used.

Basu et al. (2021)~\cite{basu2021performance} evaluated the performance of Indian LID systems on emotional speech. Speech data is collected from three Indian languages, Assamese, Bengali, and Santali. The data covered six emotional states. SDC features were used with i-vector and TDNN architectures. The same authors in~\cite{basu2021multilingual}, developed a low-resourced language (LRL) speech corpora for 16 languages of East and North-East Indian languages. It contained a total of 67.42 hours (both read speech and conversation clips) of data recorded from 240 speakers. Some languages, such as Ao, Angami, Lotha, Sumi, were previously almost zero-resourced for carrying out LID research. Further, the authors conducted comprehensive experiments with four different acoustic features, MFCC, MFCC-$\Delta$-$\Delta^2$, MDCC-SDC, and RASTA-PLP. They also used seven different classifiers, VQ, GMM, SVM, DNN, TDNN, LSTM, and i-vector-LDA. This makes a total of 28 combinations of LID systems. The authors reported LSTM architecture achieving the highest LID performance outperforming the TDNN based architectures.

Bakshi et al. (2021)~\cite{bakshi2021feature} used nine Indian languages, Assamese, Bengali, Gujarati, Hindi, Marathi, Kannada, Malayalam, Tamil, Telugu, from radio broadcasts and addressed the issue of duration mismatch between training and testing segments. They used utterance-level 1582 dimensional openSMILE feature and proposed a duration normalized feature selection (DNFS) algorithm. Experiments were validated in three different classifiers, SVM, DNN, and random forest (RF). Their proposed algorithm reduced the feature dimensions to only 150 optimized components that improved the LID accuracy among different duration mismatch test conditions. The same authors extended the approach in~\cite{bakshi2021improving} with the cascading of inter-language-family and intra-language-family based LID models. The first stage trained a binary classifier for recognizing the Indo-Aryan and Dravidian languages. The second stage trained LID models for individual languages in the families along with an additional out-of-family class. The additional class improved the overall LID accuracy even if the first stage had significant false positives. The cascaded system also helped to improve LID performance for shorter utterance duration.

In~\cite{muralikrishna2021noise}, the authors proposed LID system with nine Indian languages; Assamese, Bengali, Gujarati, Hindi, Kannada, Malayalam, Manipuri, Odia, Telugu. The proposed an utterance level representation that incorporate the frame-level local relevance factors. To capture the frame-level relevance, the authors proposed a novel segment-level matching
kernel based support vector machine (SVM) classifier. In~\cite{muralikrishna2021spoken}, the authors proposed \emph{within-sample similarity loss} (WSSL) for channel invariant language representation and used it in adversarial multi-task learning. Eight Indian languages from two different databases, the IIT-Mandi read speech corpus, and the IIITH-ILSC corpus was used. The proposed method helped reduce the adverse effects of domain mismatch and improved noise robustness. As a result, the LID performance on unseen mismatched test data improved.

Chakraborty et al. (2021)~\cite{chakraborty2021denserecognition} proposed \emph{DenseNet} architecture to build LID systems with the IITKGP-MLILSC and LDC2017S14 corpora. Using mel-spectrogram features, independent LID systems were trained for both the corpus. Their proposed architectures for both corpora outperformed several other recent neural architectures; CNN, ResNet, CNN-Bi-LSTM. The proposed architecture contained layers with multiple blocks of fully-connected segments. The input for each block was formed by concatenating the preceding block outputs. Such connections improved gradient flow for larger networks.

In our another study~\cite{dey2021cross}, the generalization capabilities of the standalone LID systems, trained with a single corpus, was investigated by cross-corpora evaluation. The three most widely used corpora: IIITH-ILSC, LDC2017S14, and IITKGP-MLILSC, were considered in this study. Five languages, Bengali, Hindi, Punjabi, Tamil, and Urdu, that are common to all three corpora, were used. Using the training data of each corpus, three standalone LID models were trained. 20-dimensional MFCC features with several feature-compensation techniques, CMS, CMVN, FW, RASTA, PCEN, were used as inputs. We used the TDNN architecture~\cite{snyder2018spoken} to train each LID model. The study revealed that even if a LID model was achieving state-of-the-art LID performance on same-corpora test utterances, its LID performance with the cross-corpora test utterances was near to the chance level. The applied feature-compensation methods, however, were shown to reduce the corpora mismatch by removing channel and background noise effects. It helped to improve the cross-corpora generalization.

Ramesh et al. (2021)~\cite{ramesh2021self} explores self-supervised approaches to build phonotactic based LID systems for seven Indian languages: Bengali, Hindi, Malayalam, Marathi, Punjabi, Tamil, and
Telugu. The audios were collected from video streaming websites. This work deals with the issue of lack of phonetically transcribed corpora for Indian languages. The authors used 512 dimensional contextualized embedding from the \emph{wav2vec} self-supervised network to build LID model with convolution RNN (CRNN) architecture. The wav2vec model had an encoder (maps each feature sample to a latent space) and aggregator network (maps sequence of latent samples into contextual representation). The authors also trained a supervised LID model with the ASR (trained on the Librispeech database) bottleneck features. The self-supervised approach outperforms the supervised model, especially for shorter utterance duration. However, due to larger contexts in phonotactics, both the approaches significantly outperformed the acoustic frame-level LID systems.

Tank et al. (2022)~\cite{tank2022novel}, conducted an comparison study of different machine learning based architectures using three Indian languages; English, Gujarati, and Hindi. They used MFCC feature with pitch and energy-based features and compared the LID performance using  Linear Discriminant, Gaussian Naïve Bayes, Fine Tree,  Linear SVM, KNN, and feed-forward neural network.

Biswas et al. (2022)~\cite{biswas2022automatic}, the authors proposed time-series aggregation of the MFCC feature for capturing language discriminating information at both macro-level and micro-level. The authors then applied the  FRESH (FeatuRe Extraction based on Scalable Hypothesis
tests)  based feature selection algorithm to reduce the complexity. The authors conducted independent LID experiments using the IIT-M and IIITH-ILSC data. For experimenting the efficiency of the proposed on global languages, VoxForge corpus was also used for experiment. The proposed feature showed efficient LID performance for the all the databases using only a shallow neural network model.

\subsection{Summary of the reviewed research works}
We summarize the key methodological details for the research works discussed above. The summary includes the used corpus, extracted features, and trained classifiers in a tabular format.
\begin{scriptsize}
\begin{longtable}[!tp]{|l|c|p{3.4cm}|p{3cm}|p{3cm}|}
\caption{Summary of methodologies followed in the major research contributions for the Indian LID systems.}
\vspace{.05cm}
\label{tab:summary}\\
\hline
\textbf{Authors} & \textbf{Year} & \textbf{Corpus} & \textbf{Feature} & \textbf{Classifier} \\ \hline \hline
\endfirsthead
\endhead
        Maity et al. \cite{maity2012iitkgp} & 2012     &    IITKGP-MLILSC      &     MFCC, LPCC    &  GMM          \\ \hline
        Verma et al. \cite{verma2013indian} &  2013    &    Self-collected (Three Indian languages)      &   MFCC      & SVM \\ \hline   
        Reddy et al.~\cite{reddy2013identification} & 2013 & IITKGP-MLILSC & MFCC with prosody & GMM \\ \hline
        Bhaskar et al. \cite{bhaskar2013analysis} & 2013 & IITKGP-MLILSC & MFCC & GMM \\ \hline
        Nandi et al. \cite{nandi2015implicit} & 2015 & IITKGP-MLILSC & MFCC, LP residual & GMM \\ \hline
        Sengupta et al. \cite{sengupta2015study} & 2015 & Self-collected (16 Indian languages) & MFCC, SFCC & GMM, SVM \\ \hline
        Kumar et al. \cite{kumar2015significance} & 2015 & IITKGP-MLILSC & MFCC & GMM-UBM \\ \hline
        Mounika et al. \cite{mounika2016investigation} & 2016 & Self-collected (12 Indian languages) & MFCC-$\Delta-\Delta^2$ & DNN with attention (DNN-WA) \\ \hline
        Sengupta et al. \cite{sengupta2016identification} & 2016 &  Self-collected (16 Indian languages) & MFCC, SFCC with SDC & GMM, SVM \\ \hline
        Bakshi et al. \cite{aarti2017spoken} & 2017 &  Self-collected (Nine Indian languages) & MFCC-$\Delta-\Delta^2$ & DNN \\ \hline
        Madhu et al.~\cite{madhu2017automatic} & 2017 & Self-collected (Seven Indian languages) & Phonotactic and prosody & DNN \\
        \hline
                Vuddagiri et al. \cite{vuddagiri2018iiith} & 2018 & IIITH-ILSC & MFCC-$\Delta-\Delta^2$ & i-vector, DNN-WA \\ \hline

                Dutta et al. \cite{dutta2018language} & 2018 & IITKGP-MLILSC & Group delay based cepstral features & GMM \\ \hline

        Veera et al. \cite{veera2018combining} & 2018 & Self-collected (13 Indian languages) & MFCC, LP residual & DNN \\ \hline

        Mukherjee et al.~\cite{mukherjee2019deep} & 2019 & IIITH-ILSC & Spectrogram & CNN \\
        \hline
        Bhanja et al. \cite{bhanja2019pre} & 2019 & NITS-LD & MFCC with prosody & GMM-UBM, i-vector, DNN \\ \hline
        Bhanja et al. \cite{bhanja2019deep} & 2019 &  NITS-LD & MFCC, MHEC & GMM-UBM, i-vector, DNN, LSTM \\ \hline
        Mandava et al. \cite{mandava2019investigation} & 2019 & IIITH-ILSC & MFCC with SDC, CTC & i-vector, TDNN, Bi-LSTM \\ \hline
        Mandava et al.~\cite{mandava2019attention} & 2019 & IIITH-ILSC & MFCC-SDC, ASR phoneme posterior & Attention based residual TDNN \\ \hline
        Das et al.~\cite{das2020bottleneck} & 2020 & Self-collected (Three Indian languages) & MFCC+SDC, LPC, bottleneck feature & Deep auto-encoder with softmax regression \\
        \hline
Paul et al.~\cite{paul2021indian} & 2020 & seven languages from IIITH-ILSC & MFCC & LSTM\\ \hline
Das et al. \cite{das2020hybrid} & 2020 & Indic TTS & BBA based fusion of MFCC and LPC & Random forest classifier \\ \hline
Bhanja et al.~\cite{china2020cascade} & 2020 & NITS-LD, OGI-MLTS & MFCC, pitch chroma, formant information & Cascaded CNN-LSTM \\ \hline
Siddhartha et al.~\cite{siddhartha2020language} & 2020 & IIITH-ILSC & Sub-band filtering on LP residual & DNN-WA \\ \hline        Deshwal et al.~\cite{deshwal2020language} & 2020 & Self-collected (four Indian languages) & Fusion of MFCC with RASTA-PLP & DNN \\ \hline
Mukherjee et al.~\cite{mukherjee2020lazy} & 2020 & self-collected (3 Indian languages) & Modified MFCC & Lazy learning based classifier\\ \hline
Das et al.~\cite{das2020bottleneck} & 2020 & Self-collected (four Indian languages) & MFCC, LPC, SDC, Bottleneck feature & Deep autoencoder \\ \hline
Garain et al.~\cite{garain2021fuzzygcp} & 2020 & IIITH and IIT Madras developed corpora & MFCC with spectral bandwidth, contrast, roll-off,
flatness, centroid, tone & DNN, CNN, and semi-supervised generative adversarial
network (SS-GAN) \\ \hline
Basu et al.~\cite{basu2021performance} & 2021 & Self-collected (Three Indian languages) & MFCC-SDC & SVM, i-vector, TDNN \\ \hline
Basu et~al.~\cite{basu2021multilingual} & 2021 & Self-collected (16 Indian languages) & MFCC with $\Delta,\Delta^2$ and SDC, RASTA-PLP & VQ, GMM, SVM, MLP, TDNN, LSTM, and i-vector-LDA \\ \hline

Bakshi et al.~\cite{bakshi2021gmm} & 2021 & Self-collected (Nine Indian languages) & MFCC-$\Delta$-$\Delta^2$, GMM supervector & GMM, DNN \\ \hline
Bakshi et~al.~\cite{bakshi2021feature} & 2021 & Self-collected (Nine Indian languages) &  Utterance-level 1582
dimensional openSMILE feature & SVM, DNN, random forest (RF) \\ \hline
Bhanja et al.~\cite{bhanja2021modelling} & 2021 & NITS-LD, OGI-MLTS & Syllable, word, and phrase level prosody and MHEC & GMM-UBM, i-vector-SVM, DNN\\ \hline
Chakraborty et~al.~\cite{chakraborty2021denserecognition} & 2021 & LDC2017S14, IITKGP-MLILSC & Mel-spectrogram & DenseNet \\ \hline
Bakshi et~al.~\cite{bakshi2021improving} & 2021 & Self-collected (Nine Indian languages) & Utterance-level 1582
dimensional openSMILE feature & Hierarchical language family based SVM, DNN, random forest (RF) \\ \hline
Muralikrishna et~al.~\cite{muralikrishna2021noise} & 2021 & Nine languages from IIITH-ILSC & BNF & Segment level matching kernel (SLMK) SVM \\ \hline
Muralikrishna et~al.~\cite{muralikrishna2021noise} & 2021 & IIIT-Mandi read speech and IIITH-ILSC & BNF & Adversarial multi-tasking with WSSL loss \\ \hline
Dey et~al.~\cite{dey2021cross} & 2021 & IIITH-ILSC, LDC2017S14, IITKGP-MLILSC & MFCC with feature-compensation & TDNN \\ \hline
Ramesh et~al.~\cite{ramesh2021self} & 2021 & Self-collected (Seven Indian languages) & wav2vec embedding, ASR BNF & CRNN \\ \hline
Tank et~al.~\cite{tank2022novel} & 2022 & Self-collected (Three Indian languages) & MFCC, pitch, and energy & LDA, KNN, SVM, decision tree, Gaussian Naive Bayes, ANN \\ \hline
Biswas et al.~\cite{biswas2022automatic} & 2022 & IIT-M, IIITH-ILSC, VoxForge & Time-series aggregated MFCC & ANN \\ \hline
\end{longtable}
\vspace{-.5cm}
\end{scriptsize}

\subsection{Analysing the present state for the Indian LID research}
\label{sec:trend}
Following our analytical review of the major research works in the field of Indian LID, we have discussed the key observations that indicate the present state-of-the-research:
\begin{itemize}
    \item Acoustic features such as MFCC with SDC or $\Delta-\Delta^2$ features are used mostly. If additional information from prosody, excitation source, phase spectrum are fused, the overall LID performance is improved. Most of the time, score-level fusion is shown to perform better as compared to feature-level fusion.
    \item From the classifier perspective, we observed that the earlier approaches until 2015 mostly used GMM or SVM models. However, gradually the trend shifts towards more complex neural networks based architectures. The recent Indian LID research works explore LSTM, TDNN, residual networks, DNN with attention models.
    \item IITKGP-MLILSC and IIITH-ILSC are the most widely used corpora. Both of these corpora covered more than 20 major Indian languages. A notable amount of the research works~\cite{sengupta2015study,aarti2017spoken,basu2021performance,deshwal2020language,mukherjee2020lazy} are conducted with self-collected \emph{in-house} corpora.
    \item The availability of speech data for each Indian language has increased with time. This trend is expected to continue because of the ease of accessing several online resources. For example, the recently developed VoxLingua107 corpus contains 14 Indian languages and at least 40 hours of speech data is available for each language.
    \item However, we find that almost all Indian LID systems are developed using speech data from broadcast news and TV shows. The pronunciations are highly professional, and the dialects and accents are standardized. In order to make the LID systems more realistic, conversational speech data should also be utilized extensively.

\end{itemize}
\vspace{-.3cm}
\subsection{Overview of other low-resourced LID systems}
\label{Sec:5.5}

In this subsection, we briefly discuss some notable LID research works in other low-resourced scenarios; Scandinavian, South African, Nigerian, and Oriental languages. The discussion about the LID research for other low-resource languages is beneficial for getting additional insights. Analyzing the different methodologies followed for low-resourced LID systems across the globe, researchers may conduct relevant adaptations to the existing Indian LID systems for performance improvements.

In~\cite{trong2019enabling}, LID systems were developed for different dialects of the \emph{North Sami} language. The authors used the under-resourced DigiSami database. The data consists of 3.26 hours of read-speech and 4.28 hours of spoken conversation. Log-Mel filterbank feature was used with $\Delta$, and $\Delta^2$ contexts, and an attention-based semi-supervised end-to-end (SSEE) system was trained with CNN, LSTM, and fully-connected architectures. The use of additional unlabeled data using semi-supervised approach compensated for the lack of training data for the proper learning of the attention weights. In~\cite{cerva2021identification}, three Scandinavian languages, Swedish, Danish, and Norwegian, were classified using ASR bottleneck features. The authors also used 39-dimensional filterbank features and trained i-vector and x-vector TDNN systems. The multilingual bottleneck feature significantly improved LID performance.

PPRLM based LID system was trained for 11 South African languages in~\cite{peche2009development}. 13-dimensional MFCC with $\Delta$ and $\Delta^2$ was used as acoustic features and fed to a context-dependent HMM model for phoneme recognition. Bi-phone statistics from each phoneme recognizer are concatenated and applied to SVM classifier for the LID task. LID system was provided ambiguous predictions for the closely related languages. Woods et al.~\cite{woods2020robust} attempted LID task with three Nigerian languages, Igbo, Yoruba, and Nigerian English. Each language contained 700 speech segments with an average duration varying from three to five seconds. Mel-spectrograms were used with an ensemble architecture of CNN and RNN.

Oriental Language Recognition (OLR) challenge has been providing speech corpora in several oriental languages~\cite{wang2016ap16,tang2017ap17,tang2019ap19}, encouraging the LID research. In~\cite{monteiro2019residual}, ten class LID task was performed with the OLR-18 challenge data. The authors used the ResNet-50 architecture and used it for multi-task learning. The two tasks learned the same set of languages with maximum likelihood setting and triplet loss minimization. In~\cite{duroselle21b_interspeech}, the authors developed a LID system that is robust against channel mismatch and noisy conditions for ten oriental languages. The system is developed by fusing three TDNN models trained with multilingual bottleneck feature with a GMM model trained on MFCC and SDC feature. The system achieved top performance in the OLR-20 challenge in its respective tasks. Kong et al.~\cite{kong21b_interspeech} proposed a dynamic multi-scale convolution-based architecture for classifying three dialects of the Mandarin Chinese language. They applied six-fold data augmentation on the OLR20 data that consists of 10 Oriental languages. 64-dimensional MFCC with three-dimensional pitch statistics features were extracted. Backend dialect recognizers were trained using the embedding extracted from the multi-scale convolution model. The system outperformed the best-performing OLR-20 submitted system for the corresponding task. 

For additional insight discussions about the closely related low-resourced languages of South Asia could have been of more significance. There are many low-resourced languages, such as Saraiki, Hindko, Brahui (Northern Dravidian language), Khowar mainly spoken in Pakistan. In Bangladesh also, in different parts different low-resourced languages are spoken. A prominent example among them is the Rohingya language along with Kurukh (Northern Dravidian language), Chakma, Tangchangya, and several others. However, in the literature, very limited research is conducted for the LID system development. The main reason for this is the lack of corpora development efforts.
\vspace{.3cm}
\section{Existing challenges \& future directions}
\label{sec:6}
In this section, we have elaborated on the existing open challenges for developing Indian LID systems. We also present the potential future research directions to mitigate these challenges.
\vspace{-.1cm}
\subsection{Issue of low-resource}
\begin{figure}[htbp]
    \centering
    \vspace{-.2cm}
    \includegraphics[trim={.6cm .1cm 1cm 2.8cm},clip,width=.85\textwidth]{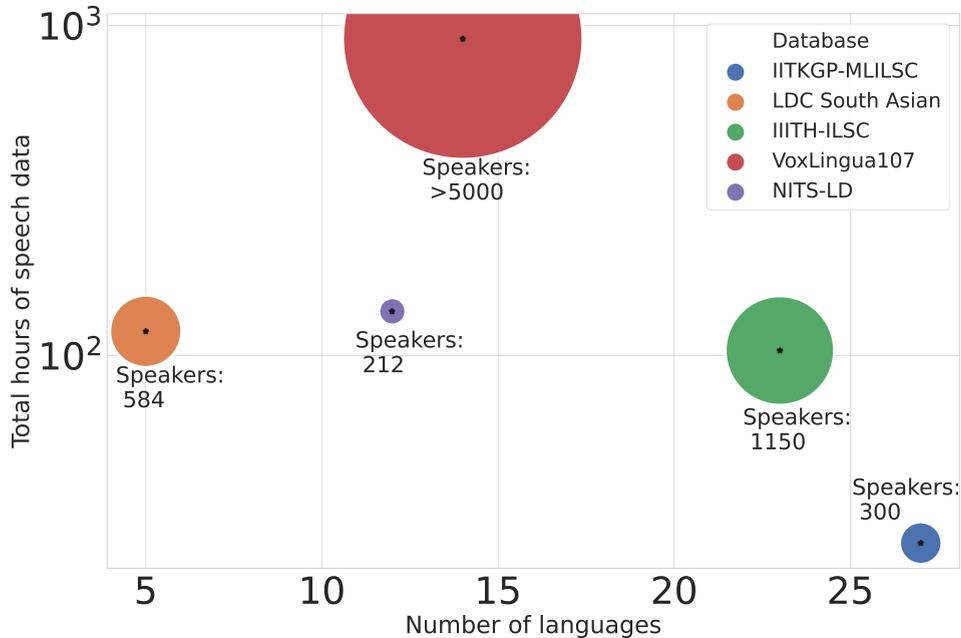}
    \vspace{-.4cm}
    \caption{Analysing the diversity of the major Indian LID corpora in terms of number of languages, total duration of data, and the total number of speakers (denoted by circle area).}
    \label{fig:scatter_size}
    \vspace{-.45cm}
\end{figure}

The low-resource problem is the most crucial factor for developing an efficient Indian LID system. Even though there are significant recent advancements in corpora development, more efforts should be made to create large-scaled diversified speech corpora for all the Indian languages. Care should be given to produce verified ground truths. The problem of low-resource becomes more critical as the state-of-the-art DNN-based architectures are data-hungry by nature and require several ten hours of audio-data per language~\cite{snyder2018spoken,plchot2018analysis}. In Fig.~\ref{fig:scatter_size}, we have graphically presented the diversities of the major Indian LID corpora in terms of number of Indian languages covered, total duration of speech data in hours, and the total number of speakers. From Fig.~\ref{fig:scatter_size}, we can observe that to mitigate the low resource issue; the developed corpora should be located in the upper right corner with a larger circular area. There are several Indian languages, for example, languages in North-Eastern India~\cite{basu2020identification} or the Northern Dravidian languages~\cite{kolipakam2018bayesian}, for which yet no significant corpora building effort has been made. Creating speech corpora for severely low-resourced languages is also an important task needed to be addressed in the future. A similar observation is also made for several other languages mainly spoken in neighbouring South Asian countries, such as Pakistan and Bangladesh. Developing speech corpora for these low-resourced languages will be of high social impact.

One potential solution for the low-resourced issue is the use of open-source corpora. LID research was primarily confined to privately developed, standardized corpora. Recently, in~\cite{arendale2020spoken}, LID system developed with three open-source corpora; \emph{Mozilla Common Voice}~\cite{mozilla}, \emph{Google Bengali speech corpus}~\footnote{\url{https://www.openslr.org/53/}}, and \emph{Audio Lingua}~\footnote{\url{https://audio-lingua.eu/?lang=fr}}, which encouraged using open-sourced speech data for LID research. Open source corpora can be developed for the Indian languages by crowd-sourcing or collecting data from the web. For each language, data should be collected from speakers from different regions, genders, age groups, and sections of the society~\cite{basu2018designing}. Variations in terms of background noise, recording channels, and room environments should be maintained~\cite{hansen2018issues}. Finally, different modes of speech, news reads, conversation, monologue should be considered. 

However, the independent open-source data collection, even from the web, can be very challenging for the rarely studied Indian languages. The amount of collected speech data can still be less. To mitigate the issue, audio data augmentation can be useful by artificially increasing the amount of speech data. For the languages with near to no standard resource available, \emph{zero-shot} learning~\cite{wang2019survey} can be an effective solution. \emph{Self-supervised learning}~\cite{ravanelli2020multi,Stafylakis2019,baevski2019vq,ramesh2021self} is also a promising approach to develop speech technologies with severely low-resourced Indian languages where verified ground-truth labels are very challenging to collect.

\subsection{Generalization of LID systems}

The issue of generalization for real-time deployment is a problem that exists in several speech processing applications. The issue becomes more severe if the amount of training data is not sufficient in volume and diversity. In Section~\ref{sec:trend}, we observe majority Indian LID systems were developed with speech from news broadcasts. The room environments in the recording studios have significantly less background noise. The recording rooms are also mostly anechoic. The pronunciations are professional, and the accent and dialects are standardized. Even due to the recording device, corpora bias can exist~\cite{dey2021cross}. Therefore, in real-world scenarios, the stand-alone Indian LID systems trained with the smaller corpus can exhibit poor generalization.

To assess the generalization issue, we have conducted a cross-corpora evaluation with the most widely used speech corpora for the Indian languages~\cite{dey2021cross}. We have shown that LID systems trained with one corpus perform poorly when the test data comes from another speech corpus. Applying several feature post-processing techniques can help to improve the generalization for cross-corpora evaluation as well as in real-world scenarios~\cite{dey2021cross}. The poor performance in the cross-corpora scenario is expected due to acoustic mismatch, session variability, speaker characteristics differences, etc. Nevertheless, generalization is an open problem which is shown by cross-corpora evaluations for other speech processing tasks as well, such as anti-spoofing~\cite{paul2017}, speech enhancement~\cite{pandey2020cross}, and speech emotion recognition~\cite{schuller2010cross}. 

Several efficient data augmentation techniques are applied on speech and speaker recognition tasks to deal with the issue of generalization. These techniques not only artificially increase the amount of data, rather increase the diversity as well~\cite{garcia2020magneto,ragni2014data}. Along with conventional speech data augmentation schemes~\cite{snyder2018x}, recently some innovative augmentation methods, such as, \emph{SpecAug}~\cite{park2019specaugment}, \emph{MixUp}~\cite{zhang2017mixup}, \emph{MicAugment}~\cite{borsos2020micaugment} are also shown to be effective for speech recognition tasks. Cross-lingual and cross-corpora generalization is improved for speech emotion recognition task using \emph{transfer learning} in~\cite{pappagari2020x,latif2018transfer}. Transfer learning is also shown to be effective for speech recognition tasks as well~\cite{wang2020improving,yi2018language}. \emph{Domain adaptation, domain generalization}~\cite{deng2014autoencoder,sun2017unsupervised}, \emph{adversarial training}~\cite{sun2018domain,fang2019channel} can also improve generalization for Indian languages by reducing the mismatch between the training and evaluation data.

\vspace{-.15cm}
\subsection{Code-switching}
\label{Sec:6_3}
India has numerous cosmopolitan cities where a large population from different cultures and languages resides in close proximity. Due to the frequent usage of different languages and mutual influence, speakers continuously keep changing the spoken language during a conversation~\cite{auer2013code}. Under the settings of the current LID systems, if any speech utterance with code-switching effect is fed as input to the LID system, the system would predict only one language class for the entire utterance even though there exists more than one languages in an alternate manner along with the temporal variations. The logical applicability of the LID predictions (along with the subsequent speech application) would be otherwise not useful, especially for the conversation speech inputs~\cite{padhi2020multilingual,lyu2013language}. Therefore, for efficient implementation in real-world scenarios, the future research direction for the Indian LID systems should also consider the \emph{language diarization}, a task that detects the instances of the change in languages in one utterance and provide the corresponding multiple predictions at the homogeneous segments. 

There are some recent research works that have explored the development of Indian LID systems from language diarization perspectives using code-switched speech utterances. In~\cite{mishra2021spoken}, the authors used synthetically code-switched data using the IIITH and L1L2 databases and applied attention-based neural networks for classification. Similarly, in~\cite{mishra2021spoken}, using the \emph{NIT-GOA Code-Switch Corpora}, language diarization was performed on code-switched Kannada-English utterances. The extracted monolingual ASR bottleneck features and trained SVM classifier for diarization. We expect the trend of incorporating language diarization in the research of LID systems will gain momentum considering the futuristic importance.

For improving code-switched ASR performance using the Indian languages, the \emph{Multilingual and Code-Switching ASR Challenges for Low Resource Indian Languages} (\emph{MUCS 2021}) challenge~\footnote{\url{https://navana-tech.github.io/MUCS2021/}} was organized. This challenge provided transcribed speech data for two code-switched pairs, Bengali-English and Hindi-English. Based on the provided data, several researchers have addressed the issue of code-switched ASR for Indian contexts~\cite{wiesner2021training, kumar21e_interspeech,sailor21_interspeech}. The \emph{Workshop on Speech Technologies
for Code-Switching in multilingual Communities} (\emph{WSTCSMC 2020}) of Microsoft research~\footnote{\url{https://www.microsoft.com/en-us/research/event/workshop-on-speech-technologies-for-code-switching-2020/}} also addressed several aspects of code-switching for spoken and textual Indian LID systems: in~\cite{rangan2020exploiting}, the authors improved the noise-robustness of code-switched LID systems with Gujarati, Telugu, and Tamil code-switched with English. They used the SpecAug augmentation method that randomly masks some time stamps and frequency channels in the input feature. In~\cite{nagarsheth2020language}, code-switched utterances from Gujarati, Telugu, and Tamil were used for the LID task. For each of the three languages, monolingual ASR models were trained. The n-grams statistics from the models are then used to train the binary LID model. Manjunath et al. (2021)~\cite{manjunath2022applications} used LID systems as a preceding stage for multilingual ASR systems in both code-switched and non-code-switched cases. The LID systems were trained with i-vector and MFCC features for different sets of Assamese, Bengali, Kannada, Odia, Telugu, and Urdu. The code-switched ASR was system then developed for the combination of Kannada and Urdu languages.

We observe that the research addressing the code-switched Indian spoken LID system is not widely explored yet. One of the key challenges for exploring code-switching in Indian LID research is lack of suitable corpora. In spite of the recent advancements, extensive efforts should be made to develop large-scale code-switching corpora for the Indian languages. So far, only few major Indian languages have code-switched standard code-switched data with English. Diverse combinations of code-switched languages can be also considered which frequently occur in daily life.

Apart from code-switching, the spoken language recognition task is also required in several other situations where dedicated application-specific processing is needed. For example, performing LID with utterances from persons with specific disabilities and diseases has an important social value~\cite{moro2020using,pulido2020alzheimer}. Similarly, performing LID with utterances from children or older people will also broaden the range of the population who can be benefited. For both the global and Indian languages, increasing the range of LID applications is an open research area.  

\vspace{-.27cm}
\subsection{Extracting language-specific features}
Several acoustic and prosodic, and phonotactic features have been explored for developing Indian LID systems. Most of the features are inspired by the efficient implementation in speech and speaker recognition tasks~\cite{dehak2011language}. For LID task, these features are also shown to perform well in the literature. However, the LID system's performance can be greatly enhanced if the signal-processing steps for the feature extraction are hand-crafted considering the linguistic characteristics of the languages to be classified. This requirement is even more essential for the Indian languages, which have a large extent of similarity and mutual influence. Ideal LID features should focus on the specific linguistic differences among the closely related languages. For example, two very related languages may have certain uniqueness in their rhythmic, tonal, and stress-related characteristics. In such cases, extraction of language-specific prosody cues can greatly benefit the recognition capability between the two languages. Phonotactics features are also can not be directly extracted from the majority of the Indian languages due to lack of phonetic transcripts. However, there are some recent multilingual ASR challenges~(discussed in Section~\ref{Sec:6_3}), which attempted to solve the issues of transcripts for the Indian languages. Alternatively, \emph{self-supervised} approaches~\cite{Stafylakis2019,baevski2019vq,ramesh2021self} and \emph{transfer learning} can also be useful to build large-scale phonotactic Indian LID systems with very limited transcripts.
\vspace{-.25cm}
\subsection{Exploration of language family information for the LID task}
Indian spoken languages have evolved from the different language families, discussed in Section~\ref{Sec:3.1}. The languages within a family are expected to share more mutual influence and similarity~\cite{sengupta2015study}. The acoustic, phonetic differences among the languages of different families can be more distinct. Therefore, a hierarchical LID system based on the different language families can help discriminate among the Indian languages. For example, some severely low-resourced North-Eastern languages do not belong to the Indo-Aryan or Dravidian language families. The front-end language family classifier can recognize these languages at the beginning, followed by a dedicated LID system. In the literature, there are few research attempts that adapted this approach~\cite{jothilakshmi2012hierarchical,sengupta2015study, sengupta2016identification, bhanja2019pre,bakshi2021improving}. Further exploration in this approach, especially with state-of-the-art neural architectures, is promising. Transfer learning based approaches can also be investigated if the requirement of the LID system is to only classify languages within a language family. First a large LID model can be trained with all the Indian languages. Then, the model can be further fine tuned by transfer learning for the specific language families. This approach is useful for the severely low-resourced language families. Quantifying the level of language similarity and mutual influence~\cite{sengupta2015study} among the different language families can also help to validate the linguistic theories. Domain and speaker robust language representations can be learned for such analysis.
\vspace{-.05cm}
\subsection{Lack of standardized experimental protocols}
Following the summarized literature discussion in Table~\ref{tab:summary}, we observe that the different research works for developing the Indian LID systems used different corpora, different sets of languages, different data-split, and different performance metrics. Due to the low-resourced issue, several in-house data are independently recorded, and experiments are conducted using the self-designed protocols. We consider that a comparative performance analysis among the different research works is fair only when the experimental protocols are standardized. Due to this issue, we also intentionally avoid presenting the numerical comparison of results for the discussed research works. To mitigate the issue, we suggest introducing a large-scale language recognition challenge dedicated to the Indian scenarios. The challenges can provide large, properly labeled development and evaluation data, fixed experimental protocols with state-of-the-art evaluation metrics. We expect that this will help systematic benchmarking of the Indian LID systems to a large extent.

\section{Conclusion}
\label{sec:7}
India is a country with a vast population that has significantly diverse cultural and linguistic backgrounds. The usage of modern voice-based smart devices and services can become a part of the daily life of Indian society. However, most of the Indian population can use these smart technologies if they can verbally communicate with the smart gadgets using their native languages. Therefore, from both social and economic perspectives, developing the Indian spoken language recognition system is an important enabler. The main challenge for developing the Indian LID systems had been the low resource. The available standard speech data for the Indian languages is not enough to efficiently implement the state-of-the-art classifier architectures. For some Indian languages, even there is hardly any effort to build standard speech corpora. Further, the Indian languages have notable resemblance and mutual influence due to their socio-political history and geographical proximity. The developed LID systems are prone to be confused among the closely related language pairs, such as Hindi-Urdu, Bengali-Odia, etc.

Due to the recent development of standard speech corpora for Indian languages, the LID research for the Indian context has been gathering momentum since the last decade. There are already significant numbers of studies for the Indian LID task. Researchers have explored various kinds of speech features and different state-of-the-art classification techniques. In this work, we have comprehensively put together all the significant research attempts for the Indian spoken language recognition field and analyzed them in detail. This review work is intended to help the researchers and future enthusiasts in this field gain an overall idea about the current state of the research progress for Indian spoken language recognition research. We have further discussed the existing open challenges for Indian LID systems along with the potential future research directions. We hope, in the future, the language recognition technology, specifically for the low-resourced Indian languages, will continue progressing in the direction that helps efficient real-time applications.

\section*{Acknowledgements}
The authors would like to thank the associate editor and the three anonymous reviewers for their careful reading, detailed comments, and constructive suggestions, which substantially enhanced the content of the manuscript.

\bibliographystyle{ieeetr}
\bibliography{sample-base}
\end{document}